\documentclass{article} 
\usepackage[preprint]{colm2026_conference}
\usepackage{microtype}
\usepackage{hyperref}
\usepackage{url}
\usepackage{booktabs}
\usepackage{graphicx}
\usepackage{amsmath}
\usepackage{enumitem}
\usepackage[font=small,labelfont=bf]{caption}
\usepackage{placeins}
\usepackage{lineno}
\usepackage{float}
\definecolor{darkblue}{rgb}{0, 0, 0.5}
\hypersetup{colorlinks=true, citecolor=darkblue, linkcolor=darkblue, urlcolor=darkblue}
\title{How LLMs Detect and Correct Their Own Errors: The Role of Internal Confidence Signals}

\author{Dharshan Kumaran\thanks{Corresponding author: dkumaran@google.com} \\
Google DeepMind \\
\And
Viorica Patraucean \\
Google DeepMind \\
\And
Simon Osindero \\
Google DeepMind \\
\And
Petar Veli\v{c}kovi\'{c} \\
Google DeepMind \\
\And
Nathaniel Daw \\
Google DeepMind \& Princeton University \\
}

\begin{document}

\ifcolmsubmission
\linenumbers
\fi

\maketitle

\section*{Abstract}
Large language models can detect their own errors and sometimes correct them without external feedback, but the underlying mechanisms remain unknown. We investigate this through the lens of second-order models of confidence from decision neuroscience. In a first-order system, confidence derives from the generation signal itself and is therefore maximal for the chosen response, precluding error detection. Second-order models posit a partially independent evaluative signal that can disagree with the committed response, providing the basis for error detection. \citet{anon2026confidence} showed that LLMs cache a confidence representation at a token immediately following the answer (i.e. post-answer newline: PANL)---that causally drives verbal confidence and dissociates from log-probabilities. Here we test whether this PANL signal extends beyond confidence to support error detection and self-correction. Here we test whether this signal supports error detection and self-correction, deriving predictions from the second-order framework. Using a verify-then-correct paradigm, we show that: (i) verbal confidence predicts error detection far beyond token log-probabilities, ruling out a first-order account; (ii) PANL activations predict error detection beyond verbal confidence itself; and (iii) PANL predicts which errors the model can correct---where all behavioural signals fail. Causal interventions confirm that PANL signals rescue error detection behavior when answer information is corrupted. All findings replicate across models (Gemma 3 27B and Qwen 2.5 7B) and tasks (TriviaQA and MNLI). These results reveal that LLMs naturally implement a second-order confidence architecture whose internal evaluative signal encodes not only whether an answer is likely wrong but whether the model has the knowledge to fix it.

\section{Introduction}
The ability of large language models to detect and sometimes correct their own errors without external feedback (e.g., \citealt{huang2024large, kamoi2024can}) is well documented but poorly understood. Confidence ratings can be obtained through diverse means---token log-probabilities \citep{kadavath2022language}, sampling-based consistency \citep{geng2023survey, tian2023just}, and verbal reports \citep{xiong2023can, yoon2025confidence, steyvers2025calibration}. In biological agents, confidence drives error monitoring, help-seeking, and strategy revision \citep{rabbitt1966errors, yeung2004neural, fleming2017self}. The self-correction literature, however, has studied \emph{whether} models can improve their answers without asking a fundamental prior question: does the model's own confidence predict whether it will identify an error and whether a correction attempt will succeed?

\begin{figure}[!t]
    \centering
    \includegraphics[width=\linewidth]{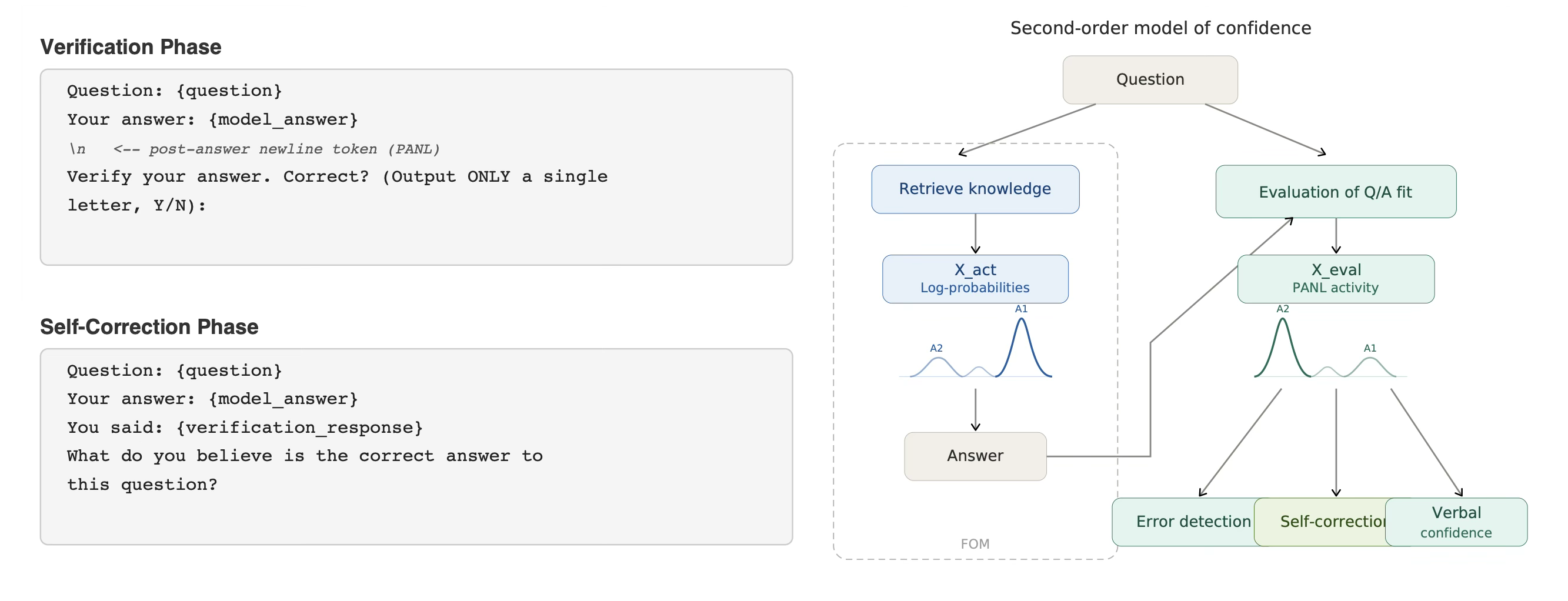}
    \caption{\textbf{Left panel}: Verification and self-correction prompt structure (see \S\ref{app:paradigm} for full details). The model's answer and verbal confidence were generated in a separate prior phase. In the verification phase, the model is shown its own answer to a TriviaQA (or MNLI) question and asked to judge whether it is correct (Y/N), followed by a self-correction prompt. Residual stream activations were extracted during the \emph{verification} phase at the post-answer newline token (PANL, indicated by arrow)---the first token after the model's answer following \citep{anon2026confidence}. \textbf{Right panel}: Second-order model of confidence, adapted from \citet{fleming2017self}. Left side (dashed box): the first-order model (FOM), in which a generation process produces an answer (here via greedy decoding) and the associated log-probabilities ($X_{\mathrm{act}}$) are the only available confidence signal. Under greedy decoding, $X_{\mathrm{act}}$ --- and therefore confidence---is by definition maximal for the chosen answer, so a purely first-order system cannot conclude it erred. Right side: the second-order extension, in which the completed answer engages a qualitatively distinct evaluative process that assesses question--answer fit by attending backward over the full response---a different computation from the retrieval process that produced it (see text for details). Because this evaluation performs a different computation over the model's knowledge, it can shift the internal distribution over possible answers such that the committed answer ($A_1$) is no longer the mode (now $A_2$). The resulting evaluative signal ($X_{\mathrm{eval}}$; termed $X_{\mathrm{conf}}$ in the original framework), encoded at answer-adjacent token positions (PANL), is partially independent of $X_{\mathrm{act}}$ and drives verbal confidence, error detection, and self-correction.}
    \label{fig:prompt_FOM_SOM}
\end{figure}

We address this question at two levels: behaviourally, we ask whether verbal confidence predicts verification and self-correction outcomes; representationally, we ask whether the model's internal activations carry richer information than its overt confidence report. \citet{anon2026confidence} showed that LLMs cache a confidence representation at the post-answer newline token (PANL) that causally drives verbal confidence and dissociates from generation-time log-probabilities. Here we ask whether this same signal extends beyond confidence to support error detection and self-correction, drawing on the distinction between first-order and second-order models of confidence from decision neuroscience \citep{fleming2017self, kepecs2008neural, kiani2009representation} (see Figure \ref{fig:prompt_FOM_SOM} and Appendix~\ref{app:related_work}). In first-order models, a single signal drives both the decision and the confidence report---the decision variable $X_{\mathrm{act}}$ determines both which response is selected and confidence in it. Applied to LLMs, the first-order analogue is token log-probabilities: under greedy decoding, confidence is by definition maximal for the chosen answer, leaving no basis for error detection. In second-order models, confidence arises from a partially independent variable $X_{\mathrm{eval}}$ that performs a qualitatively distinct evaluation of the response. Because $X_{\mathrm{eval}}$ is not yoked to $X_{\mathrm{act}}$, it can disagree with the committed response---providing the foundation for error detection \citep{fleming2017self}.

The computational difference between $X_{\mathrm{act}}$ and $X_{\mathrm{eval}}$ may be understood by analogy to the recall/recognition distinction in episodic memory \citep{brown2001recognition} (see Appendix~\ref{app:related_work}). Just as an agent can fail to recall a fact yet recognise a retrieved answer as wrong---because recall and recognition draw on partially independent substrates---$X_{\mathrm{act}}$ and $X_{\mathrm{eval}}$ perform qualitatively different computations over the same underlying knowledge. Generation proceeds via parametric fact retrieval \citep{meng2022locating}---analogous to recall---and $X_{\mathrm{act}}$ (log-probabilities) reflect this process, yoked to whichever answer was selected, whether correct or not. $X_{\mathrm{eval}}$, by contrast, operates analogously to recognition: because PANL signals are computed after the answer is complete and can attend backward over the full generated response, it can assess question--answer fit---akin to recognising whether a retrieved answer is correct, rather than performing the retrieval itself. As illustrated in Figure~\ref{fig:prompt_FOM_SOM}, this evaluative process can shift the model's internal distribution over possible answers such that the committed answer is no longer the mode, even when it was the argmax during generation. It is precisely this decoupling---between the distribution that drove generation and the distribution implicit in the post-hoc evaluation---that constitutes the second-order architecture and enables error detection and self-correction.

The second-order framework generates specific predictions that we test in this study. If the model maintains an evaluative signal at PANL that is partially independent of the generation signal, then: (i) verbal confidence should predict error detection beyond token log-probabilities, since the evaluative signal carries information that the generation signal does not; (ii) PANL activations should predict error detection beyond verbal confidence, since the model's internal evaluation is only partially externalised in its confidence report; and (iii) if the evaluative process has independent access to the model's knowledge, PANL activations should predict which errors the model can correct, where behavioural confidence signals cannot (see Figure~\ref{fig:prompt_FOM_SOM}). We test these predictions on factual question-answering rather than reasoning tasks---precisely because in factual tasks without chain-of-thought there is no reasoning trace to revisit, so successful self-correction requires the evaluative process to access the model's knowledge in a qualitatively different way from generation.

Our main contributions are:
\begin{enumerate}[leftmargin=*, itemsep=2pt, topsep=4pt]
\item \textbf{Behavioural second-order signature.} Verbal confidence predicts verification responses far beyond token log-probabilities, ruling out a first-order account---but predicts only \emph{whether} the model detects errors, not \emph{if} it can correct them.
\item \textbf{PANL predicts correctability where all behavioural signals fail.} PANL activations predict which answer revisions will succeed---where no combination of behavioural signals will---indicating that the model internally encodes information about its ability to improve that it does not communicate in its overt outputs.
\item \textbf{Causal role in error detection.} Activation patching confirms that PANL is causally sufficient to rescue error detection when answer information is corrupted; joint ablation reveals that the evaluative signal is shared across PANL and the last answer token, with PANL isolating the evaluative component from the answer representation itself.
\item \textbf{Generality across models and tasks.} All key findings replicate in Gemma 3 27B and Qwen 2.5 7B, using the TriviaQA and MNLI datasets, suggesting a domain-general evaluative mechanism.
\item \textbf{Second-order confidence architecture.} Together, these results provide the first evidence that LLMs spontaneously implement a second-order confidence architecture in the sense of \citet{fleming2017self}: an evaluative process that is partially independent of generation, can disagree with the committed response, and encodes structured information about the model's ability to improve.
\end{enumerate}
A direct practical implication follows: monitoring PANL activations could enable selective self-correction in base models, and may reflect the very mechanism reasoning models exploit to trigger backtracking~\citep{ward2025reasoning, gandhi2025cognitive, yang2025stepback}---intervening only when the model's own representations indicate it has the knowledge to produce a better answer.

\section{Related Work}
Here we briefly describe previous work that is most clearly related to ours (please see Appendix~\ref{app:related_work} for a detailed review of the literature). LLMs have been shown to have variable capacities to detect their own errors and self-correct (termed intrinsic self correction)\citep{huang2024large, kamoi2024can, zhang2025understanding, madaan2023self, liu2024large, chen2025sets, weng2023large}. \citet{li2024confidence} use a prompt instructing the model that if it is very confident in its answer then it should maintain it, otherwise it should update. However, their analysis does not examine whether confidence signals predict error detection, nor whether such signals predict downstream self-correction beyond behavioral output measures. Bertolazzi et al. (2025) \citep{bertolazzi2025validation} use circuit analysis to study arithmetic error detection in small LLMs (1.5B–3B), finding that models rely on "consistency heads" — attention heads performing surface-level digit matching. Their work identifies the responsible circuit components -- rather than the nature of representations involved -- and does not study the role of confidence. 

A growing body of work demonstrates that LLMs encode information about the quality of their outputs within internal activations, often in ways that diverge from surface-level confidence \citep{azaria2023internal, liu2025llm, burns2022discovering, li2023inference, burger2024truth, orgad2024llms, lu2025unified}. 

\citet{anon2026confidence} showed that verbal confidence is not computed on-the-fly when a rating is requested. Instead, the model automatically caches a confidence representation at the first token following the generated answer---a post-answer newline token (PANL) -- which is later readout as a verbal confidence. Because the transformer's causal attention mask means PANL can attend to all question and answer tokens but not to any subsequent prompt tokens (i.e. concerning verification instruction, or confidence reporting; see Methods), the PANL residual stream state constitutes a backward-looking summary of the completed response. Convergent causal evidence established that PANL activations at middle layers in LLMs play a causal role in the generation of verbal confidence, and that they explain substantial variance in verbal confidence \emph{beyond} answer token log-probabilities---indicating a richer evaluation of question--answer fit rather than a simple readout of generation fluency. The present paper asks whether this same signal supports error detection and self-correction.

\section{Experiments}
We study Gemma 3 27B \citep{team2025gemma} as our primary model on the TriviaQA factual knowledge dataset \citep{joshi2017triviaqa} ($n = 7{,}227$ questions), with cross-model replication on Qwen 2.5 7B ($n = 3{,}500$) and cross-task replication on MNLI \citep{williams2018broad}. All answers were generated with greedy decoding (temperature $= 0$), ensuring that A1 represents the model's argmax of its token distribution. Any improvement from A1 to A2 therefore cannot arise from stochastic resampling; it requires the evaluative process at PANL to access a different weighting over alternatives in which the committed answer is no longer dominant (Figure~\ref{fig:prompt_FOM_SOM}), enabling the model to revise toward a response it did not initially select. Full model and dataset details are in \S\ref{app:models}.

In the simple verify-then-self-correct paradigm we employ, the model generates an answer and reports its confidence (Phase~0), is then shown its own answer and asked to verify it (``Y''/``N''; Phase~1 (verification)), and finally produces a second answer (Phase~2 (self correction); see \S\ref{app:paradigm}; Figure~\ref{fig:prompt_FOM_SOM}). We extract residual stream activations at PANL during the verification phase and use linear probes to predict verification behaviour and second-attempt correctness. We first characterise the predictive value of behavioural signals and then turn to the activation analyses to ask whether PANL representations explain variance that behavioural signals leave unaccounted for. 

\subsection{Basic Behavioral Results}
Gemma 3 27B scored 75.5\% on the TriviaQA dataset at its first attempt (A1: $n = 7{,}227$ questions). Verbal confidence and the log-probabilities of the answer tokens yielded an AUROC of 0.74 and 0.76 for predicting A1 correctness, respectively; expected calibration error (ECE) was 0.13 and 0.20, respectively. Self-correction yielded a modest but highly significant improvement in accuracy, from 75.5\% at A1 to 79.2\% at its second attempt (A2; $\Delta = +3.7$ percentage points; McNemar's $\chi^2 = 202.3$, $p < .001$).

\begin{figure}[!t]
    \centering
    \includegraphics[width=0.6\linewidth]{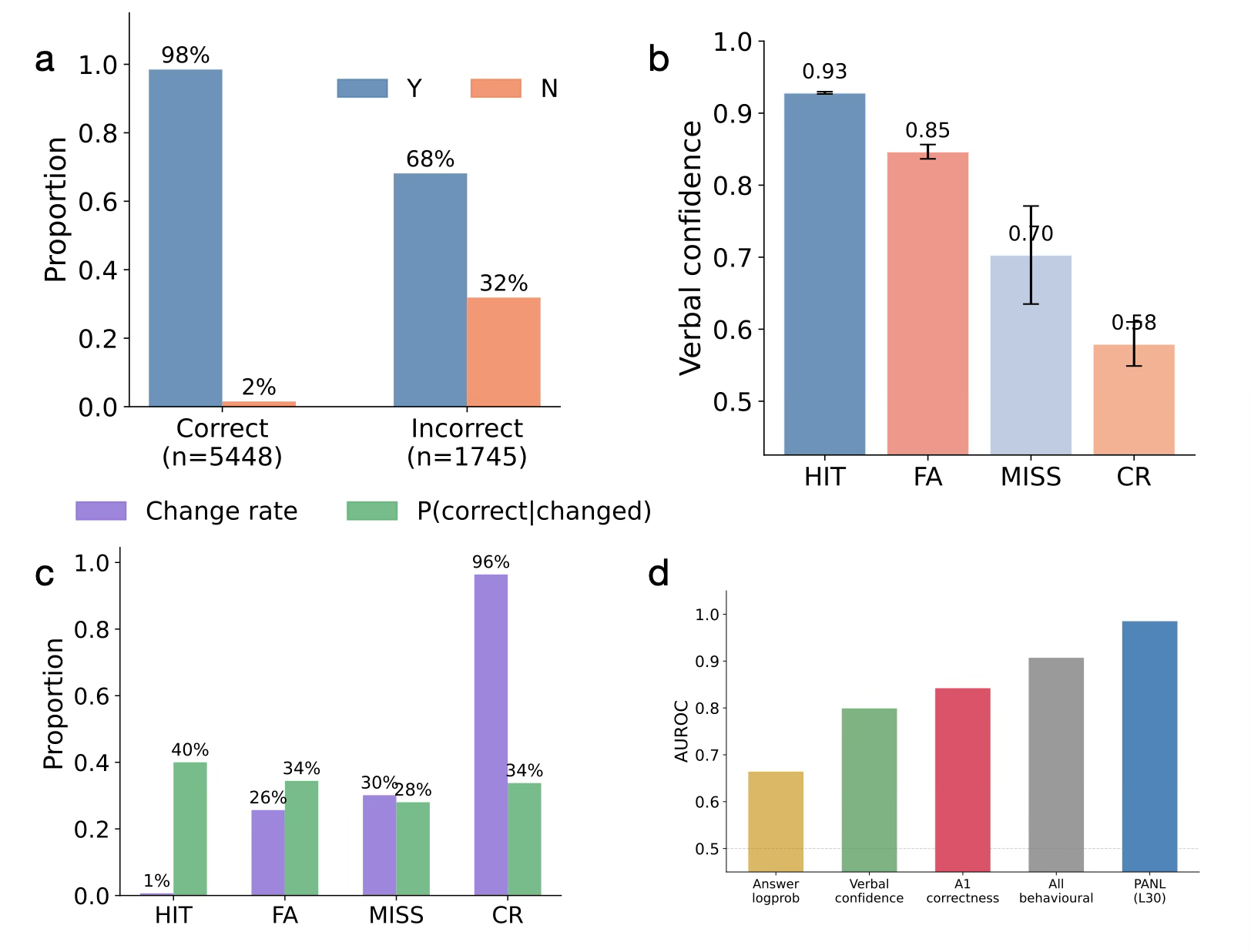}
    \caption{Gemma 3 27B behavioural results on TriviaQA ($n = 7{,}193$). \textbf{a}: Verification responses by A1 correctness. The model displays robust error detection ($d' = 1.67$) with a strong Y-bias ($c = -1.34$): it confirms 98\% of correct answers but still endorses 68\% of incorrect ones. \textbf{b}: Mean verbal confidence ($\pm$95\% CI) across SDT cells (e.g.\ Hit = A1-correct given a Y response at verification (i.e.\ V=Y). FA: A1-incorrect given V=Y). \textbf{c}: Answer change rate (purple) (i.e. proportion of answers changed between A1 and A2) -- and correction success conditional on changing (green) by SDT cell. Verification gates whether the model revises, but correction success is roughly constant ($\sim$28--40\%) regardless of cell. \textbf{d}: AUROC for predicting the verification response from answer token log-probabilities, verbal confidence, A1 correctness, their combination and PANL activation at layer 30. Verbal confidence substantially outperforms log-probabilities, consistent with a second-order evaluative signal that is partially independent of generation fluency. PANL outperforms best behavioral baseline.}
    \label{fig:gemma_behavioral_auroc} 
\end{figure}

\paragraph{Prediction 1: Gemma shows a reliable error detection capacity}
Figure~\ref{fig:gemma_behavioral_auroc} shows verification responses as a function of A1 correctness. The model displays robust error-detection sensitivity ($d' = 1.67$) with a strongly conservative criterion ($c = -1.34$), reflecting a pronounced Y-bias. A foil experiment confirms this is not a fixed property: when the model verifies answers it did not generate, error detection ability scales with foil plausibility ($d' = 2.57$ for hard foils to $5.08$ for unrelated foils) and the Y-bias diminishes accordingly ($c = -0.78$ to $-0.14$; see \S\ref{app:foil_method}, \S\ref{app:foil}, and Figure~\ref{fig:foil_composite}).

\paragraph{Prediction 2: Verbal confidence carries predictive information beyond log-probabilities for predicting verification behavior.}
A second-order account predicts that verbal confidence, because it is driven by an evaluative signal partially independent of the generation process, should predict the verification response beyond what token log-probabilities alone can explain. Note that although a reliable verbal confidence signal should by definition predict binary correctness of the initial response (as it does, AUROC = 0.74 for A1 correctness reported above), it need not necessarily predict verification response. 

We tested the role of confidence in predicting the verification response (V=Y vs V=N) with nested logistic regressions . A model containing only mean answer log-probability yielded AUROC $= .668$. Adding verbal confidence produced a dramatic improvement (AUROC $= .832$; LR~$\chi^2 = 839.4$, $p < 10^{-184}$), with standardised coefficients confirming that confidence dominates ($\beta_{\mathrm{conf}} = +0.51$, $\beta_{\mathrm{logprob}} = +0.10$). Adding verbal confidence to a model already containing log-probability \emph{and} A1 correctness yielded a similarly large improvement (AUROC: $.766 \to .865$; LR~$\chi^2 = 629.0$, $p < 10^{-138}$). The pattern is even clearer within incorrect trials, where log-probabilities are entirely uninformative about the verification response (AUROC $= .481$) while verbal confidence alone achieves AUROC $= .737$ (LR~$\chi^2 = 164.6$, $p < 10^{-37}$); log-probabilities add nothing once confidence is included ($\chi^2 = 0.37$, $p = .54$). Verbal confidence thus carries substantial evaluative information about the model's own answer that generation-time log-probabilities do not capture (see Figure~\ref{fig:gemma_behavioral_auroc}D). This is consistent with verbal confidence reflecting an evaluative process that is at least partially independent from generation fluency (i.e. token log-probabilities) ---the second-order signal identified mechanistically in \citet{anon2026confidence}.

\begin{figure}[!t]
    \centering
    \includegraphics[width=0.6\linewidth]{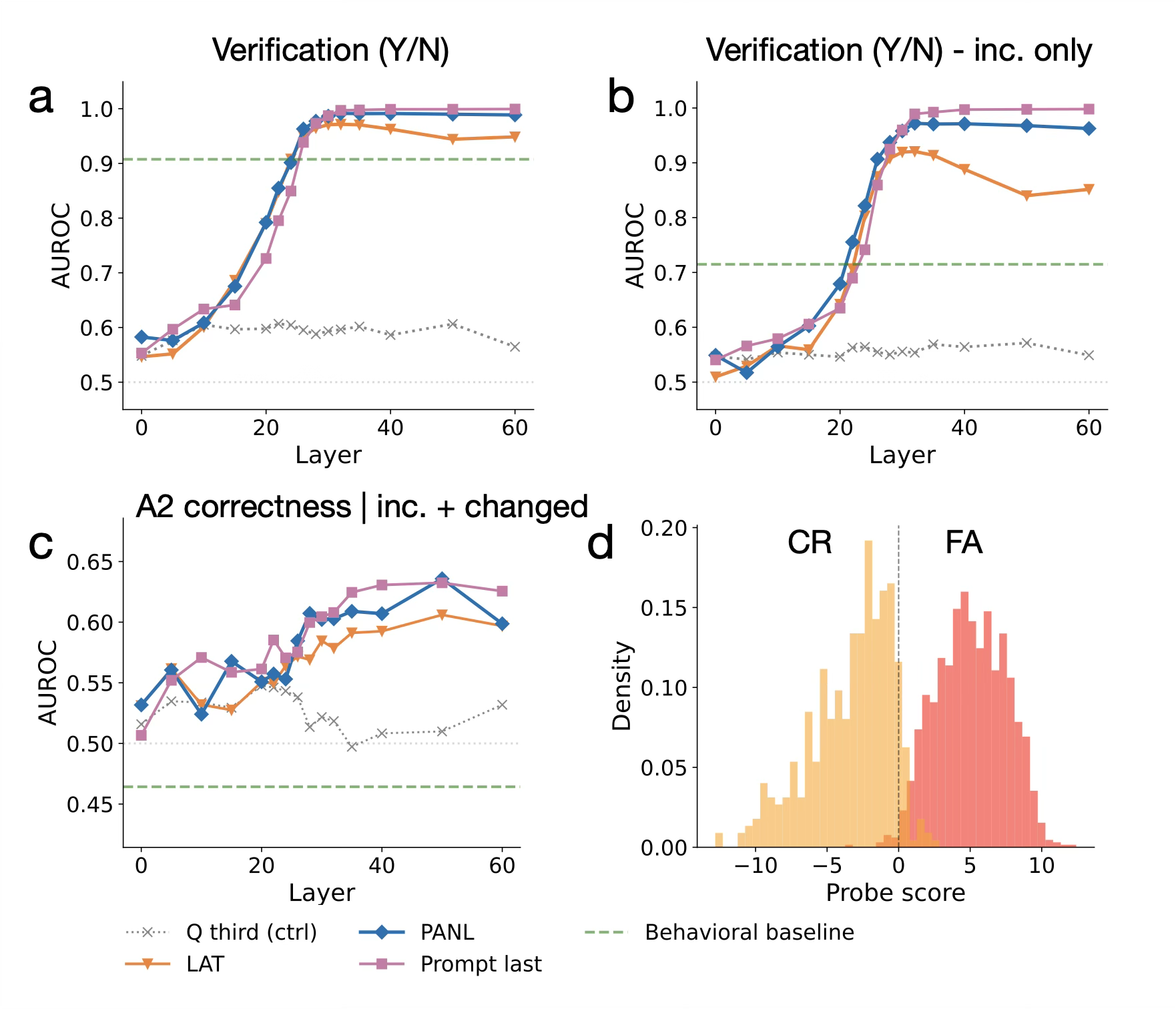}
    \caption{Linear probing results across layers for three targets: 
    \textbf{A}: verification response (all trials), 
    \textbf{B}: verification response (incorrect trials only), 
    \textbf{C}: A2 correctness (incorrect trials that changed answer). 
    Probes are $L_2$-regularised logistic regression ($C = 0.001$, 5-fold CV) on residual stream activations at four token positions: question third token (control), last answer token (LAT), post-answer newline (PANL), and prompt last token. The dashed green line indicates the best behavioural baseline (see Table~\ref{tab:AUROC_main_summary}). PANL exceeds the behavioural baseline across all three targets at mid-to-upper layers, with the largest margin for A2 correctness among changed incorrect trials (C), where behavioural predictors are at chance. The control position shows no improvement for any target. 
    \textbf{D}: Distribution of PANL L30 probe scores (trained on verification Y/N) within incorrect trials, split by verification response. The probe cleanly separates correct rejections (CR; model detected its error) from false alarms (FA; model missed its error), with minimal overlap near the decision boundary (AUROC $= .892$). This signal is encoded at PANL during the answer phase, before the verification instruction is presented.}
    \label{fig:probe_main}
\end{figure}

\begin{figure}[t]
\centering
\includegraphics[width=0.7\linewidth]{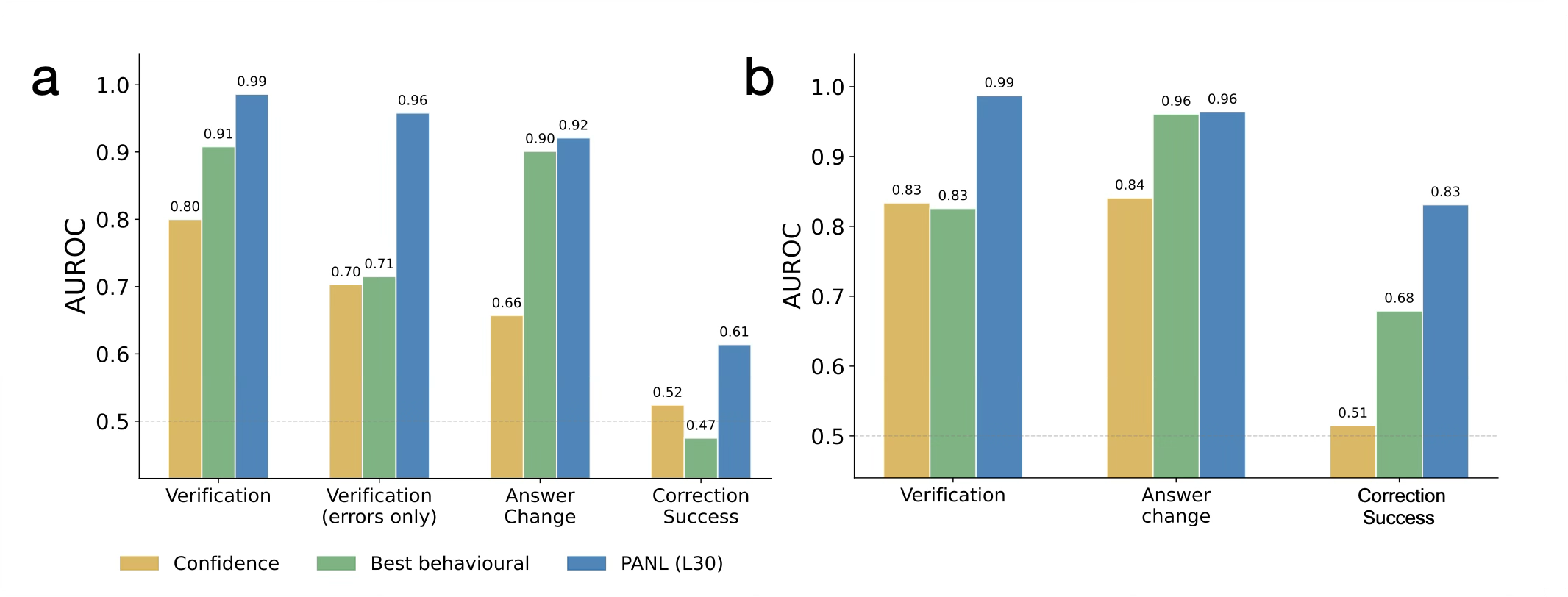}
\caption{\textbf{a}) Main experiment: AUROC for predicting verification response (all trials), verification response (incorrect trials only), answer change (incorrect trials), and A2 correctness (incorrect trials that changed answer). \textit{Verbal confidence}: the model's stated confidence from Phase~0. \textit{Best behavioural baseline}: logistic regression over available behavioural scalars---answer log-probability, verbal confidence, and A1 correctness for verification; answer log-probability, verbal confidence, and verification log-probability difference for answer change and A2 correctness. \textit{PANL probe}: linear probe on post-answer newline activations at layer~30. The PANL probe matches or exceeds the best behavioural baseline across all targets, and is the only predictor substantially above chance for A2 correctness among changed trials (AUROC~$= .614$ vs.\ $.475$ for the behavioural baseline; see Table~\ref{tab:AUROC_main_summary} for summary; Tables~\ref{tab:supp_verif}--\ref{tab:supp_correct} for individual predictor breakdowns). \textbf{b}) Analogous results for the foil experiment, averaged across hard, easy, and unrelated foil conditions. Verification within incorrect trials is not shown because the candidate answer is always incorrect by construction. See Table~\ref{tab:foil_auroc_full} for per-condition results.}
\label{fig:auroc_barchart_main}
\end{figure}

\paragraph{Prediction 3: PANL activations predict verification behavior better than verbal confidence.}
To test whether PANL representations carry evaluative information beyond what behavioural signals capture, we trained linear probes on residual stream activations at the PANL position (see \S\ref{app:probing}) and extracted a cross-validated \emph{probe score}---a scalar compression of the high-dimensional activation that can be compared directly with behavioural predictors. Figure~\ref{fig:probe_main}A shows AUROC across layers for four token positions. PANL activations at mid-to-upper layers substantially exceed the behavioural baseline; the control position (third question token) remains at baseline throughout. The last prompt token shows comparable predictive power, as do downstream positions (PANL+1, offset 6--18; Figure~\ref{fig:tqa_linear_probe_suppl}A)---the former unsurprisingly, since it directly precedes the Y/N verification decision; the latter carry decodable but not causally relevant information, as we show in \S\ref{app:causal_detail}. The advantage is even more pronounced within incorrect trials (Figure~\ref{fig:probe_main}B), where the verification decision is most consequential. At layer~30, PANL achieves AUROC $= .986$ against the best behavioural baseline of $.908$ (LR~$\chi^2 = 1100.0$, $p < .001$; Table~\ref{tab:AUROC_main_summary}; Figure~\ref{fig:auroc_barchart_main}A). The advantage is far larger within incorrect trials---where the verification decision is most consequential and behavioural signals are weakest---with PANL achieving AUROC $= .958$ against a baseline of just $.715$ (Figure~\ref{fig:auroc_barchart_main}A and Table~\ref{tab:supp_verif}). This confirms P3: the model's evaluative representation contains more information about its own accuracy than it communicates in its overt confidence report---verbal confidence is a partial, lossy readout of a richer internal signal. 

Figure~\ref{fig:probe_main}D illustrates this separation within incorrect trials: the PANL L30 probe score cleanly distinguishes errors that the model will flag (correct rejections) rather than accept as correct (false alarms; AUROC $= .892$), with minimal overlap---indicating that the evaluative signal discriminating detected from undetected errors is already encoded at PANL---before the verification instruction itself.

\paragraph{The evaluative signal predicts answer revision.}
The model changes its answer far more often when it reports an error (V=N) than when it endorses its answer (V=Y): within incorrect trials, 97\% of correct rejections led to an answer change compared with 27\% of false alarms (Figure~\ref{fig:gemma_behavioral_auroc}C). This pattern tracks verbal confidence: correct rejections have the lowest mean confidence across SDT cells and the highest change rate, while false alarms have substantially higher confidence and rarely trigger revision (Figure~\ref{fig:gemma_behavioral_auroc}B,C). Within incorrect trials, the verification logprob difference---a continuous measure of the strength of the verification decision---is the strongest single behavioural predictor of answer change (AUROC $= .907$; Table~\ref{tab:supp_change}), consistent with answer revision being driven by the same evaluative signal that produces the verification response. PANL improves on even this strong baseline (AUROC $= .921$ vs $.901$ combined behavioural; LR~$\chi^2 = 237.7$, $p < .001$; Table~\ref{tab:AUROC_main_summary}; Figure~\ref{fig:auroc_barchart_main}A), indicating that the internal evaluative representation carries finer-grained information about whether the model will revise than any of its behavioural outputs capture.

\paragraph{Prediction 4: PANL predicts which errors the model can correct, where behavioural signals fail.}
The preceding analyses show that the evaluative signal at PANL predicts both whether the model detects an error and whether it revises its answer. But the most consequential question for self-correction is whether the model can predict which revisions will \emph{succeed}. The behavioural evidence suggests not: conditional on changing its answer, the model produces a correct A2 approximately one-third of the time regardless of SDT cell ($\sim$29--34\%; Figure~\ref{fig:gemma_behavioral_auroc}C), and neither verbal confidence (AUROC $= .524$) nor the verification logprob difference (AUROC $= .531$) predicts correction success within incorrect trials that changed their answer (Table~\ref{tab:supp_correct}). The combined behavioural baseline falls below chance (AUROC $= .475$), confirming that no combination of the model's overt signals distinguishes correctable from uncorrectable errors once the decision to revise has been made.

Note that we focus on incorrect trials that changed their answer ($n = 856$) because this is the only subset where predicting correctability is non-trivial: correct trials are almost exclusively hits that do not change, so A2 correctness across all trials is dominated by the trivial prediction that unchanged correct answers remain correct.

PANL activations succeed where behavioural signals fail. The PANL probe score predicts correction success with AUROC $= .614$ (LR~$\chi^2 = 9.5$, $p < .01$; Table~\ref{tab:AUROC_main_summary}; Figure~\ref{fig:auroc_barchart_main}A), and this advantage is visible across mid-to-upper layers at the PANL position (Figure~\ref{fig:probe_main}C). The foil experiment, and the experiment using the MNLI dataset (Appendix~\ref{sec:mnli_probing}), provides evidence to bolster this finding: a PANL probe predicts correction success with AUROC $.813$--$.858$ across foil types---substantially higher than for own-answer trials (Figure~\ref{fig:auroc_barchart_main}B). This likely reflects the larger and more varied pool of answer-change trials in foil conditions, where the model rejects foil answers far more often than its own, providing greater variance for the probe to exploit. Behavioural predictors remain at or near chance for correction success across all foil conditions (Table~\ref{tab:foil_auroc_full}).These findings confirms P4: the model's internal representation at PANL encodes information about whether it has the knowledge to produce a better answer---information that is not captured by any behavioural signal, including the verification decision itself. Within the second-order framework, this is the signature of an evaluative process that has access to the model's knowledge structure: the same process that assesses question--answer fit also provides evidence about what the correct answer should be.

This result holds even after controlling for the model's latent knowledge. Adding P(IK)---a probe-derived estimate of retrieval reliability for each question \citep{kadavath2022language}---to the behavioural baseline does not diminish PANL's contribution to predicting either answer change (LR~$\chi^2 = 101.7$, $p < 10^{-23}$) or A2 correctness (LR~$\chi^2 = 99.0$, $p < 10^{-23}$). The verification and P(IK) probe weight vectors are near-orthogonal at layer~30 (cosine similarity $= +0.007$), confirming that the error-detection signal is geometrically independent of the model's latent knowledge state rather than a proxy for question difficulty (see \S\ref{app:pik} for details).

\paragraph{Additional analyses}
The PANL probe score correlates $r = 0.91$ with the verification logprob difference and $r = 0.60$ with verbal confidence, but only $r = 0.21$ with answer log-probability ($r = 0.02$ within incorrect trials), consistent with PANL encoding a pre-computed (i.e. before the verification instruction) evaluative judgment that is partially independent of generation-time fluency (\S\ref{app:signal_char}). Additional probing analyses---including FA-specific answer change and correctability breakdowns, continuous prediction of verification strength, and information propagation to downstream token positions---are reported in \S\ref{app:probing_suppl} (see also Figure~\ref{fig:tqa_linear_probe_suppl}).

\paragraph{Summary of behavioral and linear probing analyses.}
Together, these results confirm all four predictions of the second-order framework. Verbal confidence carries evaluative information beyond log-probabilities (P2); PANL activations predict verification beyond all behavioural signals, particularly within incorrect trials where the gap is largest (P3); and PANL predicts which errors the model can correct, where behavioural signals fail entirely (P4). The evaluative signal at PANL is thus richer than what the model externalises in its overt confidence report, or its confidence in the verification decision (as indexed by the log-probability difference between Y/N) ---encoding not only whether the answer is likely wrong, but whether the model has the knowledge to fix it. We next ask whether this signal plays a causal role in error detection.

\section{Causal Interventions}
To establish whether the evaluative information at PANL plays a causal role in error detection, we performed activation patching and noising experiments on Gemma 3 27B ($n = 1{,}000$ TriviaQA trials; see \S\ref{app:patching} and \S\ref{app:noising} for methods; see \S\ref{sec:qwen_causal} for Qwen 2.5 7B results).

\begin{figure*}[!t]
    \centering
    \includegraphics[width=0.6\linewidth]{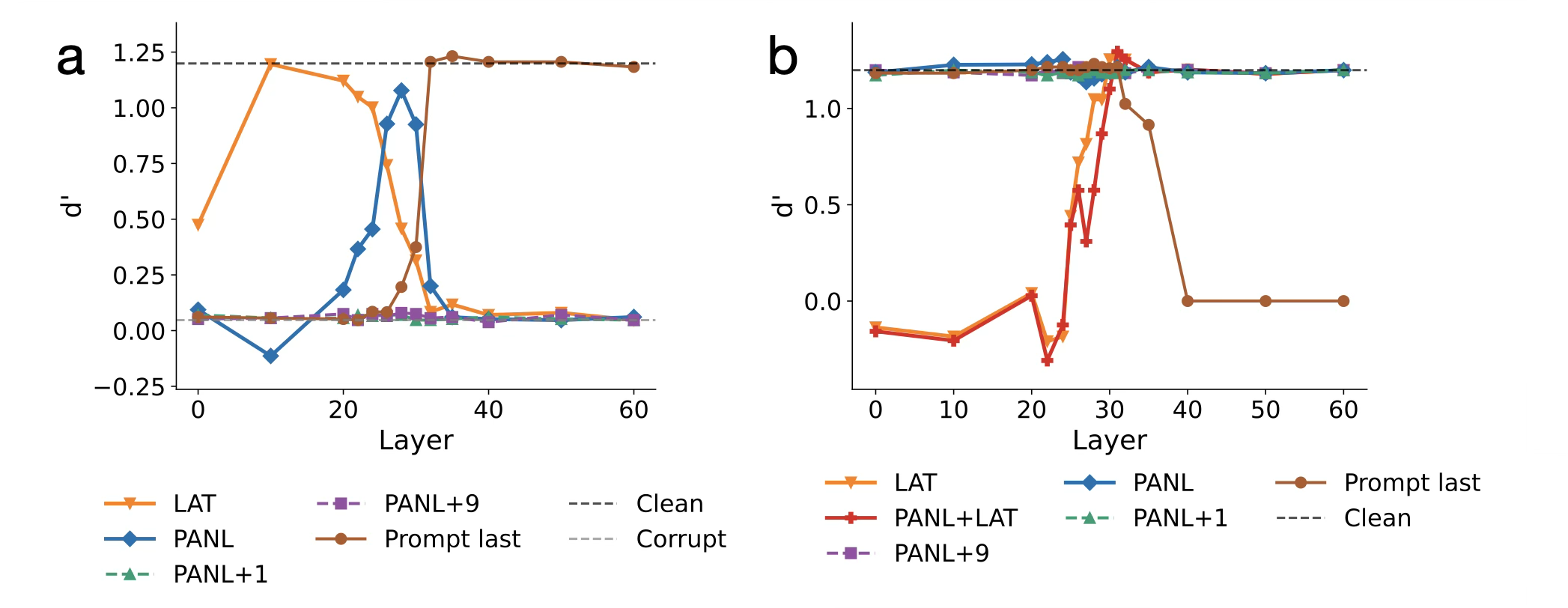}
    \caption{Gemma 3 27B causal experiments on TriviaQA ($n = 1{,}000$). \textbf{a}: Activation patching. Clean $d' = 1.17$; corrupted $d' = 0.09$ (grey dashed). LAT (orange) rescues at early-to-mid layers (peak $d' = 1.23$ at L15, $\sim$100\% recovery); PANL (blue) rescues at mid layers (peak $d' = 0.89$ at L30, $\sim$74\% recovery); prompt last token (brown) rescues at later layers (peak $d' = 1.25$ at L35). PANL+1 and PANL offset 9 show no rescue despite carrying decodable information. \textbf{b}: Activation noising through mean ablation. Ablating LAT alone (orange) severely disrupts error detection at early-to-mid layers ($d' = -0.21$ at L22), recovering by L30. Ablating PANL alone (blue) has no effect. Jointly ablating LAT and PANL (red) produces additional deficits at mid layers where LAT is recovering (L27: $d' = 0.31$ vs $0.82$ for LAT alone), consistent with representational redundancy. Ablating the prompt last token (brown) disrupts at later layers ($d' = 0.0$ at L40--60).}
    \label{fig:gemma_causal}
\end{figure*}

\paragraph{PANL is causally sufficient for error detection.}
We corrupted the verification prompt by replacing all answer token embeddings with position-specific means, abolishing error detection ($d'$: $1.17 \to 0.09$). Restoring clean activations at a single position and layer reveals three causally relevant positions with complementary temporal profiles (Figure~\ref{fig:gemma_causal}A): LAT rescues at early-to-mid layers ($\sim$100\% recovery at L15), PANL at mid layers ($\sim$74\% recovery at L30), and the prompt last token at later layers ($\sim$107\% recovery at L35). Positions where linear probes decode verification-predictive information (PANL+1, PANL offset~9) show no rescue at any layer, confirming that decodable information is not always causally expressed in behaviour. The temporal dissociation between LAT and PANL is consistent with an evaluative computation that begins at the answer tokens and is consolidated at PANL in mid-to-upper layers---the same layers where probing performance peaks.

\paragraph{PANL is not necessary, but carries the evaluative signal redundantly with LAT.}
Mean ablation---replacing a position's activation with its layer-specific mean from a balanced calibration set---reveals a complementary pattern (Figure~\ref{fig:gemma_causal}B). Ablating PANL alone has no effect at any layer ($|\Delta d'| \leq 0.062$), while ablating LAT severely disrupts error detection at early-to-mid layers ($d' = -0.21$ at L22), recovering by L30. However, jointly ablating LAT and PANL produces deficits beyond LAT alone at mid layers (e.g., $d' = 0.31$ vs $0.82$ at L27), precisely where PANL's probing performance peaks. This additional deficit is only visible when LAT is also disrupted---consistent with the two positions carrying the evaluative signal redundantly. Note that LAT confounds the answer representation with any evaluative signal computed over it; PANL provides a cleaner window onto the evaluative computation (see \S\ref{app:causal_detail} for full details; Figure~\ref{fig:gemma_causal} for effect of activation patching/noising on the strength of the verification decision (i.e. Y/N log-proability difference)).

\paragraph{Cross-model and cross-task generalisation.}
Qwen 2.5 7B replicates all key findings (see \S\ref{sec:qwen_probing} and \S\ref{sec:qwen_causal} for full results; Figures~\ref{fig:qwen_probing} and \ref{fig:qwen_causal}): robust error detection ($d' = 1.63$), PANL predicting verification, answer change, and correction success beyond behavioural baselines (AUROC $= .774$ for A2 correctness within incorrect trials, baseline $.568$), and the same causal architecture---PANL sufficient, LAT necessary, both redundant at mid layers. To test whether the key findings are specific to factual QA---in particular, the unique ability of PANL to predict the success of self-correction---we replicated the paradigm on the MNLI dataset (which involves labelling statements as entailment, contradiction, or neutral), restricting analyses to neutral trials where the model displays meaningful verification variability (see \S\ref{app:mnli_behav} and \S\ref{sec:mnli_probing} for full results; Figure~\ref{fig:mnli_behav_composite}). On MNLI neutral trials, despite substantially weaker behavioural signals, PANL replicates all probing results including correction success within incorrect changed trials (AUROC $= .862$, baseline at chance). Cross-task probe transfer reveals that the error-detection signal generalises functionally, though correctability directions are task-specific (cosine $\leq .027$; Table~\ref{tab:cross_transfer}; see \S\ref{sec:transfer}).

\section{Conclusion}
\citet{anon2026confidence} showed that LLMs cache an evaluative representation at PANL that drives verbal confidence and dissociates from log-probabilities. We show that this second-order signal extends to error detection and self-correction---and, critically, predicts which errors the model can correct, where all behavioural signals fail. The second-order framework from decision neuroscience \citep{fleming2017self} explains why: a first-order system cannot conclude its own output is wrong, because the signal that selected the answer is by definition peaked at that answer. We hypothesize that this automatic error signal may be precisely what reasoning models have learned to act on: redeployed at intermediate commitment points within a reasoning trace to trigger backtracking~\citep{ward2025reasoning, gandhi2025cognitive, yang2025stepback}, and potentially offering a principled source of dense intermediate reward for reasoning training.

\clearpage
\section*{Acknowledgements}
We thank Leonidas Guibas and Andrea Banino for comments on an earlier version of this manuscript. We also acknowledge the use of Gemini for assistance with coding and improving the clarity of the writing.
\bibliography{refs_ED}
\bibliographystyle{colm2026_conference}

\appendix
\clearpage
\appendix
\setcounter{figure}{0}
\setcounter{table}{0}
\renewcommand{\thefigure}{A\arabic{figure}}
\renewcommand{\thetable}{A\arabic{table}}
\section*{Appendix Overview}
\label{app:overview}
\begin{itemize}
    \item \textbf{Appendix A: Methods} (\S\ref{app:methods})
    \begin{itemize}
        \item A.1 Models and Datasets (\S\ref{app:models})
        \item A.2 Experimental Paradigm (\S\ref{app:paradigm})
        \item A.3 P(IK): Probability that I Know (\S\ref{app:pik_method})
        \item A.4 Linear Probing (\S\ref{app:probing})
        \item A.5 Length-Normalised Answer Log-Probability (\S\ref{app:logprob})
        \item A.6 Activation Patching (\S\ref{app:patching})
        \item A.7 Activation Mean Ablation (\S\ref{app:noising})
    \end{itemize}
    \item \textbf{Appendix B: Detailed Review of Literature} (\S\ref{app:related_work}) — Detailed review of self-correction, latent representations, mechanistic interpretability, and decision neuroscience literature
    \item \textbf{Appendix C: Supplemental Figures} (\S\ref{app:supp_figures}) — Prompts, supplemental probing figures, causal logit difference figures
    \item \textbf{Appendix D: Supplemental Tables} (\S\ref{app:tables}) — Individual predictor AUROCs for verification, answer change, and correctability (Tables~\ref{tab:supp_verif}--\ref{tab:cross_transfer})
    \item \textbf{Appendix E: Supplemental Results --- Gemma 3 27B on TriviaQA}
    \begin{itemize}
        \item E.1 Error Detection is Graded by Answer Plausibility (\S\ref{app:foil}) — Foil experiment behavioural results
        \item E.2 PANL Signal Characterisation (\S\ref{app:signal_char}) — Correlations between probe score and behavioural variables
        \item E.3 P(IK) Orthogonality (\S\ref{app:pik}) — Independence of error-detection and knowledge signals
        \item E.4 Additional Probing Analyses (\S\ref{app:probing_suppl}) — FA/CR subsets, continuous verification prediction, downstream propagation
        \item E.5 Causal Experiment Details (\S\ref{app:causal_detail}) — Full patching and ablation results, joint ablation, LAT confound
    \end{itemize}
    \item \textbf{Appendix F: Cross-Model and Cross-Task Generalisation}
    \begin{itemize}
        \item F.1 MNLI Behavioural Results (\S\ref{app:mnli_behav})
        \item F.2 MNLI Probing Results (\S\ref{sec:mnli_probing})
        \item F.3 Cross-Task Transfer (\S\ref{sec:transfer})
        \item F.4 Qwen 2.5 7B Behavioural Results (\S\ref{app:qwen_behav})
        \item F.5 Qwen 2.5 7B Probing Results (\S\ref{sec:qwen_probing})
        \item F.6 Qwen 2.5 7B Causal Experiments (\S\ref{sec:qwen_causal})
    \end{itemize}
\end{itemize}

\clearpage

\section{Methods}
\label{app:methods}

\subsection{Models and Datasets}
\label{app:models}

\paragraph{Models.}
We study two instruction-tuned language models: Gemma 3 27B (\texttt{google/gemma-3-27b-it}; 62 layers, 5376-dimensional residual stream) and Qwen 2.5 7B (\texttt{Qwen/Qwen2.5-7B-Instruct}; 28 layers, 3584-dimensional residual stream). Both models were accessed via HuggingFace Transformers and run with greedy decoding (temperature $= 0$) throughout.

\paragraph{Datasets.}
We use TriviaQA \citep{joshi2017triviaqa} as our primary dataset, a factual question-answering benchmark requiring retrieval of real-world knowledge. After deduplication, our TriviaQA sample comprises 7,227 questions for Gemma and 3,500 for Qwen. We additionally test on the Multi-Genre Natural Language Inference (MNLI) dataset \citep{williams2018broad}, a three-way classification task (entailment, contradiction, neutral), using the development set downloaded from HuggingFace ($n = 9{,}888$) and applied to Gemma only. As entailment and contradiction trials show near-ceiling verification endorsement rates with minimal variance, all MNLI analyses are restricted to neutral trials ($n = 3{,}395$).

\subsection{Experimental Paradigm}
\label{app:paradigm}

\begin{figure}[H]
  \centering
  \includegraphics[width=0.9\linewidth,height=0.6\textheight,keepaspectratio]{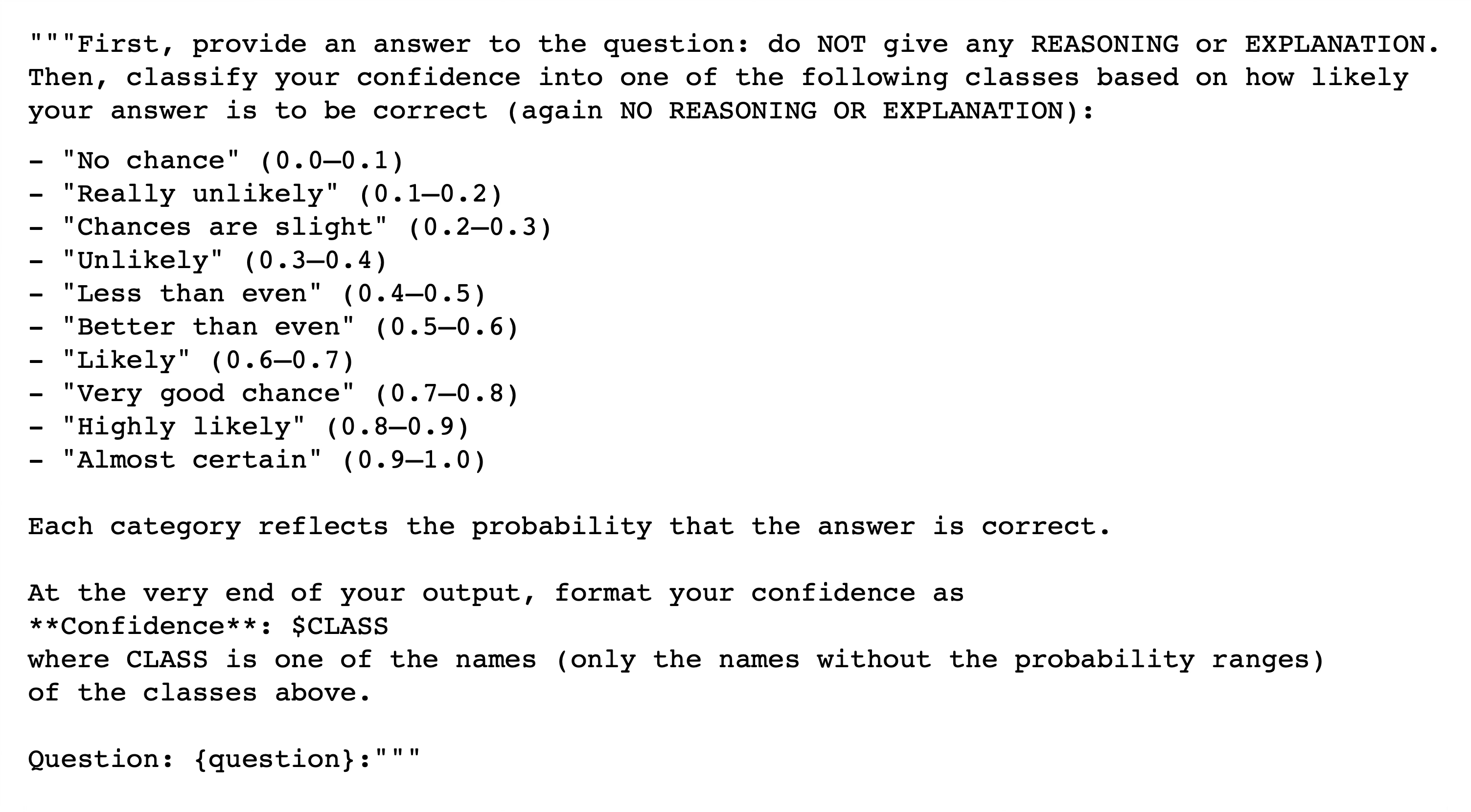}
  \caption{Full categorical confidence class prompt used in Phase 0. This prompt is derived from \citep{yoon2025reasoning} but with the following key modification: the first token of every confidence class is unique, allowing us to meaningfully analyze changes in the ID of first token, the logit of the first token etc.}
  \label{fig:conf_prompt}
\end{figure}

\paragraph{Phase 0: Answer generation and confidence rating.}
In Phase~0, the model was presented with questions and asked to generate a concise answer followed by a verbal confidence rating (see Figure~\ref{fig:conf_prompt} for the full prompt and class definitions). The prompt instructed the model to select one of ten confidence classes spanning the $[0, 1]$ interval (e.g., ``No chance'' for $0.0$--$0.1$, ``Almost certain'' for $0.9$--$1.0$). The model was explicitly instructed not to provide reasoning or explanation, yielding short answers typically one to three tokens in length. For MNLI, the same Phase~0 prompt structure was used, with the model generating a classification label (entailment, neutral, or contradiction) and a confidence rating.

\paragraph{Phase 1: Verification.}
In Phase~1, the model was presented with its own Phase~0 answer and asked to verify its correctness. The model was shown the original question and its prior answer and prompted to judge whether the answer was correct, outputting only a single letter (``Y'' or ``N'') (see Figure~\ref{fig:prompt_FOM_SOM}). The verification prompt contained the model's answer followed by a newline token (NL) and then the verification instruction:

\begin{quote}
\small
\texttt{Question: \{question\}} \\
\texttt{Your answer: \{model\_answer\}} \\
\texttt{$\langle$NL$\rangle$} \hfill $\leftarrow$ \textit{PANL: activations extracted here} \\
\texttt{Verify your answer. Correct? (Output ONLY a single letter, Y/N):}
\end{quote}

\noindent The newline token (NL) immediately following the model's answer is the position we term PANL (post-answer newline). Due to causal attention, PANL can attend to all question and answer tokens but not to any subsequent tokens in the verification instruction. It therefore has the potential to reflect a backward-looking evaluative signal relating to question--answer fit. Residual stream activations were extracted at PANL following the methodology established in \citet{anon2026confidence}.

For MNLI, the Phase~1 prompt followed the same structure:
\begin{quote}
\small
\texttt{Question: \{question\}} \\
\texttt{Your answer: \{model\_answer\}} \\
\texttt{Verify your answer. Correct? (Output ONLY a single letter, Y/N):}
\end{quote}

\paragraph{Phase 2: Self-correction.}
In Phase~2, the model received the original question, its prior answer, and its own verification response, and was asked to provide what it believed to be the correct answer:

\begin{quote}
\small
\texttt{Question: \{question\}} \\
\texttt{Your answer: \{model\_answer\}} \\
\texttt{You said: \{verification\_response\}} \\
\texttt{What do you believe is the correct answer to this question?}
\end{quote}

\noindent For MNLI, the Phase~2 prompt was adapted to the classification task:
\begin{quote}
\small
\texttt{Question: \{question\}} \\
\texttt{Your answer: \{model\_answer\}} \\
\texttt{Verify your answer. Correct?: \{verification\_response\}} \\
\texttt{What is the correct relationship - entailment, neutral, or contradiction?} \\
\texttt{(Output ONLY one of: entailment, neutral, contradiction):}
\end{quote}

\noindent We denote the Phase~0 answer as A1 and the Phase~2 answer as A2. The verification response partitions trials into four signal-detection cells: hits (correct answer, verified ``Y''), misses (correct answer, verified ``N''), correct rejections (incorrect answer, verified ``N''), and false alarms (incorrect answer, verified ``Y'').

\paragraph{Answer scoring.}
Because A2 responses were typically discursive---often containing reasoning, hedging, or restated context rather than a concise answer---we used GPT-4o to judge the correctness of both A1 and A2 responses against the ground-truth TriviaQA answer. For consistency, GPT-4o was used to score A1 as well, despite A1 answers being short enough for string matching in principle. We verified that GPT-4o scoring of A1 (and in fact A2) was highly consistent with exact string matching (agreement rate 93.2\% and 83.8\% in A1 and A2, respectively), confirming that the use of a neural judge did not introduce systematic bias.

\subsubsection{TriviaQA: Foil Experiment}
\label{app:foil_method}
To test whether the model's error-detection capacity extends beyond self-generated answers, we constructed a foil verification experiment using a subset of 1,929 TriviaQA questions. For each question, we used GPT-4o to generate three types of foil answers: a hard foil (a plausible but incorrect alternative, semantically close to the correct answer and in the same category or domain), an easy foil (a topically related but clearly wrong answer that anyone with basic knowledge should reject), and an unrelated foil (an answer from a completely different domain that is immediately obvious as incorrect). The model was also presented with its own Phase~0 answer as a baseline condition. Each question appeared in all four conditions, enabling within-question comparisons. The verification prompt was identical to the main experiment except that ``Your answer'' was replaced with ``The candidate's answer'', removing any indication that the answer was self-generated. Self-correction followed the same procedure as the main experiment. We assessed error-detection sensitivity ($d'$), response criterion ($c$), verbal confidence, answer change rate, and correction success for each condition.

\subsection{P(IK): Probability that I Know}
\label{app:pik_method}
Following \citet{kadavath2022language}, we estimated the model's latent knowledge about each question using sampling consistency. For each question, we generated 20 independent responses at temperature $= 1$ and computed the proportion that matched the ground-truth answer, yielding a question-level estimate of retrieval reliability (P(IK)) with 21 discrete levels ($0/20$ through $20/20$). P(IK) was included in nested regression analyses to ensure that PANL's contribution is not reducible to question-level knowledge.

\subsection{Linear Probing}
\label{app:probing}
Following \citet{anon2026confidence}, we trained $L_2$-regularised logistic regression probes ($C = 0.001$, 5-fold stratified CV) on $z$-scored residual stream activations from all 7{,}227 TriviaQA trials. We extracted activations at four positions: PANL (the first token following the answer), the last answer token (LAT), the prompt's last token, and a control position (third question token). Binary targets were: (i)~verification response, (ii)~answer change, and (iii)~A2 correctness; for the continuous target of verification log-probability difference we used Ridge regression and report Pearson~$r$. To isolate the contribution of PANL beyond behavioural signals, we compare against a baseline logistic regression trained on A1 correctness, mean answer log-probability, verbal confidence, and verification log-probability difference. To obtain a scalar summary of PANL information for use in downstream analyses, we extracted each trial's cross-validated predicted probability (i.e., the prediction from the fold in which it was held out); we refer to this as the \emph{probe score} and enter it as a predictor alongside behavioural measures in logistic regressions and AUROC comparisons.

\paragraph{Surface feature controls for linear probing.}
To rule out the possibility that linear probes exploit shallow textual statistics rather than genuine internal representations, we constructed a surface-feature baseline comprising TF-IDF lexical features (100 features each for question and answer text, fit separately) and token lengths of both question and answer. A logistic regression trained on these 202 surface features achieved near-chance prediction of verification response (AUROC $= .564$), answer change ($.564$), and A2 correctness ($.585$), and added no predictive value when appended to the behavioural scalar baseline ($p = 1.0$ for all targets). This confirms that the probing results reported above cannot be attributed to surface properties of the input or output text.

\subsection{Length-Normalised Answer Log-Probability}
\label{app:logprob}
To obtain a white-box measure of model confidence, we extracted token-level log-probabilities during answer generation. For each generated sequence, we computed the log-probability of each token $t_i$ given the preceding context:
\begin{equation}
    \log p(t_i \mid t_{<i}, \mathbf{x})
\end{equation}
where $\mathbf{x}$ denotes the input prompt. We then computed the mean log-probability over the $n$ answer tokens:
\begin{equation}
    \bar{\ell} = \frac{1}{n} \sum_{i=1}^{n} \log p(t_i \mid t_{<i}, \mathbf{x})
\end{equation}
This length-normalised metric controls for variation in answer length, ensuring that longer answers are not penalised simply for having more tokens. Answer token boundaries were identified by mapping the extracted answer string back to the generated token sequence.

\subsection{Activation Patching}
\label{app:patching}

\paragraph{Corruption of answer tokens via mean ablation.}
To test whether specific position--layer combinations are causally sufficient for error detection, we use a corrupt-and-restore procedure following \citet{meng2022locating, heimersheim2024use, zhang2023towards, wang2023interpretability}. We first disrupted the model's access to answer information through mean ablation of answer tokens. Let $\mathbf{x}_i^{(0)}$ denote the embedding of token $i$ in the input sequence, and let $\mathcal{A} = \{a_1, \ldots, a_k\}$ denote the set of answer token positions in the verification prompt. We computed mean activations from a calibration set $\mathcal{C}$ of 100 trials (disjoint from the test set):
\begin{equation}
\bar{\mathbf{x}}_j^{(0)} = \frac{1}{|\mathcal{C}|} \sum_{c \in \mathcal{C}} \mathbf{x}_{j,c}^{(0)}
\end{equation}
\noindent where $\mathbf{x}_{j,c}^{(0)}$ is the embedding at answer position $j$ for calibration trial $c$. For each test trial, we replaced all answer token embeddings with their corresponding mean activations, propagating this corruption through the entire forward pass. This mean ablation effectively destroys the correspondence between the question and the specific answer presented for verification, replacing answer-specific information with a position-averaged signal that retains no trial-level content.

\paragraph{Patching procedure.}
Let $\mathbf{h}_p^{(\ell)}$ denote the residual stream activation at position $p$ after layer $\ell$. For a given test trial, let $\mathbf{h}_p^{(\ell, \text{clean})}$ denote the activation from a clean forward pass (no corruption) and $\mathbf{h}_p^{(\ell, \text{corrupt})}$ denote the activation when answer tokens have been mean-ablated. Our patching intervention selectively restores the clean activation at position $p$ and layer $\ell$:
\begin{equation}
\mathbf{h}_p^{(\ell, \text{patched})} = \mathbf{h}_p^{(\ell, \text{clean})}
\end{equation}
\noindent while all other positions retain their corrupted activations. This intervention was applied after the MLP block at each layer. We tested five positions: the last answer token (LAT), PANL (post-answer newline), PANL+1, PANL offset~9, and the last token of the verification prompt. We evaluated rescue of error detection using two measures: $d'$ (sensitivity to errors) and the verification logprob difference ($\log P(\text{Y}) - \log P(\text{N})$).

\subsection{Activation Mean Ablation}
\label{app:noising}

To test whether representations at specific token positions are \textit{necessary} for error detection, we performed mean ablation experiments. For each position of interest, we replaced its residual stream activation at a target layer with the mean activation computed from a balanced calibration set of 200 trials (50 per SDT cell: hit, miss, false alarm, correct rejection). This calibration set was disjoint from the test set. Balancing across SDT cells ensures that the mean activation is not biased toward any particular verification outcome, so the replacement removes trial-specific information without injecting a systematic signal.

We note that mean ablation does not push the model toward a semantically meaningful ``neutral'' verification state. The mean of activations encoding correct and incorrect evaluations is not itself an encoding of intermediate confidence---analogous to how averaging word embeddings for ``brilliant'' and ``terrible'' does not yield a representation of ``mediocre.'' Rather, mean ablation disrupts the position's contribution to the computation by replacing trial-specific information with an uninformative average.

We tested five positions ($n = 1{,}000$ trials per layer per position): the last answer token (LAT), PANL (post-answer newline), PANL+1, PANL offset~9, and the last token of the verification prompt. For each position, we applied mean ablation at individual layers across the network and measured the resulting disruption to verification behaviour. We assessed disruption using two measures: $d'$ (error-detection sensitivity) and the verification logprob difference ($\log P(\text{Y}) - \log P(\text{N})$). If a position is necessary for error detection, ablating it should reduce $d'$ toward chance and shift the logprob difference. If ablation has no effect, the position is not necessary for the model's error-detection mechanism.
\clearpage

\section{Detailed Review of the Literature}
\label{app:related_work}

\paragraph{Self-Correction in LLMs.}
\citet{huang2024large} define intrinsic self-correction as when an LLM tries to correct errors using its inherent abilities. Self-correction more generally has wider definitions including provision of external feedback (extrinsic self-correction) --- reviewed by \citet{kamoi2024can}. Huang et al.\ show using a particular prompt that LLMs can't correct reasoning errors and can even cause performance decreases. \citet{zhang2025understanding} ask models ``are you sure, think again'' and show that performance decreases. Self-Refine \citep{madaan2023self} iteratively refines the output of the model until verified as correct, showing improvement, as do \citet{liu2024large, chen2025sets}. Self-Enhanced Test-Time Scaling (SETS; \citealt{chen2025sets}) combines parallel sampling with iterative self-verification and self-correction using a single LLM to show increased performance in reasoning tasks. \citet{weng2023large} show that self-verification of reasoning enhances performance. Self-verification is a diagnostic step where the model judges whether a proposed solution satisfies the task constraints, producing a verdict. Self-correction is the subsequent generative step where, conditioned on that verdict and the history of prior attempts, the model produces an improved solution.

\citet{li2024confidence} look at \emph{intrinsic} self-correction. They use an IoE prompt --- if you are very confident in your answer maintain it, otherwise update. They show small gains over a standard prompt. \citet{yang2025confidence} show that fine-tuning can improve self-correction, but this could be because the model learns a high-confidence--stay and low-confidence--switch policy. \citet{xie2026know} extract normalized confidence scores from self-evaluation token probabilities (Yes/No) and show that SFT preserves calibration while RL and DPO induce overconfidence through reward exploitation. However, their analysis does not examine whether internal representations encode confidence signals prior to self-evaluation, nor whether such signals predict downstream self-correction beyond behavioral output measures.

\citet{bertolazzi2025validation} use circuit analysis to study arithmetic error detection in small LLMs (1.5B--3B), finding that models rely on ``consistency heads'' --- attention heads performing surface-level digit matching --- and that validation circuits fire before arithmetic computation completes in higher layers. Their work studies artificially introduced errors on a constrained task and identifies the responsible circuit components. Our approach differs in studying naturally occurring errors on open-domain factual QA, using probes to characterise what \emph{information} the residual stream encodes rather than which \emph{components} comprise the circuit.

\paragraph{Latent Representations of Uncertainty and Correctness.}
A growing body of work demonstrates that LLMs encode information about the quality of their outputs within internal activations, often in ways that diverge from surface-level confidence. \citet{azaria2023internal} showed that models encode a latent representation of truthfulness in their hidden states: classifiers trained on these activations distinguish true from false statements with 60--80\% accuracy, generalizing across topics and outperforming prompting-based methods. Critically, classifiers trained on these internal activations outperform methods based on output probabilities, which are confounded by factors such as sentence length and token frequency and thus provide less reliable signals of truthfulness. Similarly, \citet{liu2025llm} found that activations at the first output token predict answer correctness with approximately 75\% accuracy, and introduced metrics to distinguish correct, incorrect, and irrelevant retrieval contexts directly from model internals. These findings align with broader evidence that task-relevant information may be decodable from network activations yet remain dissociated from observed behavior \citep{burns2022discovering, li2023inference, burger2024truth, lu2025unified}.

\citet{orgad2024llms} probe LLM hidden states to decode output correctness, finding that the truthfulness signal is concentrated at exact answer tokens --- consistent with our finding that token position matters critically for error-related decoding. They taxonomise errors by sampling consistency (e.g., consistently wrong vs.\ occasionally wrong) and show these types are predictable from internal representations, revealing that models can encode the correct answer while consistently generating an incorrect one. Our work differs in probing a graded confidence signal within a verification paradigm and linking it to error detection and self-correction, rather than decoding binary correctness.

\paragraph{Mechanistic Interpretability: Activation Steering.}
Activation steering is a technique for causally intervening on model behavior by adding or subtracting directions in activation space. \citet{turner2023steering} demonstrated that abstract concepts such as love or hate are encoded as approximately linear directions in transformer representations. These directions can be recovered through various methods, including contrasting mean activations between conditions that differ along a conceptual dimension. Subsequent work has applied steering to modify instruction-following behavior \citep{stolfo2024improving}, reasoning \citep{venhoff2025base}, sycophancy and other traits \citep{panickssery2023steering}, and evaluation-aware responses \citep{hua2025steering}.

\paragraph{Activation Patching.}
Activation Patching (also termed causal tracing) identifies which model components are causally responsible for specific behaviors. \citet{meng2022locating} used the corrupt-then-restore paradigm: inputs are corrupted to disrupt model output, then clean activations are selectively restored at specific position-layer combinations to measure recovery. Positions that restore performance when patched are implicated as causally sufficient for the computation. This approach has been extended to study indirect object identification \citep{wang2023interpretability}, and best practices for activation patching metrics have been systematically evaluated \citep{zhang2023towards, heimersheim2024use}.

\paragraph{Theories of Confidence in Decision Neuroscience.}
Our investigation connects to a longstanding debate in decision neuroscience concerning the computational basis of confidence \citep{fleming2017self, pouget2016confidence, kepecs2012computational, kiani2009representation}. Under first-order accounts, confidence arises from the same internal signals that drive the decision itself. In perceptual tasks, for instance, both the choice and confidence derive from a single decision variable representing accumulated evidence; confidence is simply a readout of how strongly this variable favored the chosen option. Translated to LLMs, a first-order account would hold that verbal confidence is a readout of token log-probabilities---the same signals that determined which answer tokens to generate also determine confidence.

Under second-order accounts, confidence involves signals that are distinct from---though correlated with---those driving the decision \citep{fleming2017self}. These additional signals enable an evaluation of the decision that goes beyond the information directly used to produce it. For LLMs, evidence that verbal confidence reflects information beyond token log-probabilities would suggest a second-order-like computation capable of genuine answer-quality evaluation.

An important consequence is that second-order architectures can support error detection: because confidence draws on partially independent information, the system can recognize that a response may be wrong even after committing to it. In contrast, pure first-order architectures cannot detect errors, because confidence and decision accuracy are yoked to the same underlying signal.

\paragraph{Recall, Recognition, and the Computational Basis of Error Detection}
\label{app:recall_recognition}
The distinction between generation-time signals and the evaluative signal encoded at PANL has a natural analogue in the memory literature. In episodic memory, recall and recognition dissociate both behaviourally and neurally: recall requires active reconstruction of a memory trace from a retrieval cue, whereas recognition can proceed via familiarity signals that do not depend on successful recollection \citep{brown2001recognition}. An agent can fail to recall something it can nonetheless recognise as wrong or unfamiliar, because the two processes draw on partially independent computational substrates.

The mechanistic basis for why generation-time signals in LLMs are analogous to recall is provided by work on factual retrieval. \citet{meng2022locating} demonstrated that factual associations are localised in MLP weight matrices at specific layers, with answer generation corresponding to a key-value lookup in which the subject token representation activates a stored association to produce the predicted output. On this view, token log-probabilities index the strength of parametric retrieval---how cleanly the subject-to-fact association fired---rather than its accuracy. A confidently retrieved but incorrect association yields high log-probability and, in a purely first-order system, provides no signal that an error has occurred.

The PANL representation is structurally positioned to perform a different computation. As established in \citet{anon2026confidence}, PANL attends backward over the completed answer and encodes a summary of question--answer fit that explains substantial variance in verbal confidence beyond token log-probabilities. This backward-looking integration over a finished object---rather than a forward lookup through parametric memory---is precisely what distinguishes recognition from recall: the evaluative process is not hostage to whether retrieval succeeded, because it operates on the output of retrieval rather than performing it. 

Taken together, the \citeauthor{meng2022locating} mechanistic account and the \citeauthor{brown2001recognition} computational framework provide complementary explanations for why generation-time and evaluative signals dissociate. The former specifies what generation consists of in LLMs; the latter provides the principled computational reason why a post-hoc evaluation over the completed response should have access to information that the generation signal does not.

\clearpage

\section{Supplemental Figures}
\label{app:supp_figures}

\begin{figure}[htbp]
    \centering
    \includegraphics[width=\linewidth]{figures/Gemma_probing_suppl_composite.png}
    \caption{Supplementary linear probe results. Increase in AUROC (delta) above behavioral baseline shown for PANL and several downstream positions. \textbf{A}: Verification prediction from PANL, first answer token, last answer token, and downstream positions (offset 6, 12, 18 tokens after PANL). \textbf{B}: Predicting A2 correctness within correct rejection (CR) trials. \textbf{C}: Predicting A2 correctness within false alarm (FA) trials. \textbf{D}: Prediction of verification confidence (logprob\_diff) as a continuous variable ($\Delta$ Pearson $r$).}
    \label{fig:tqa_linear_probe_suppl}
\end{figure}

\begin{figure}[!t]
    \centering
    \includegraphics[width=\linewidth]{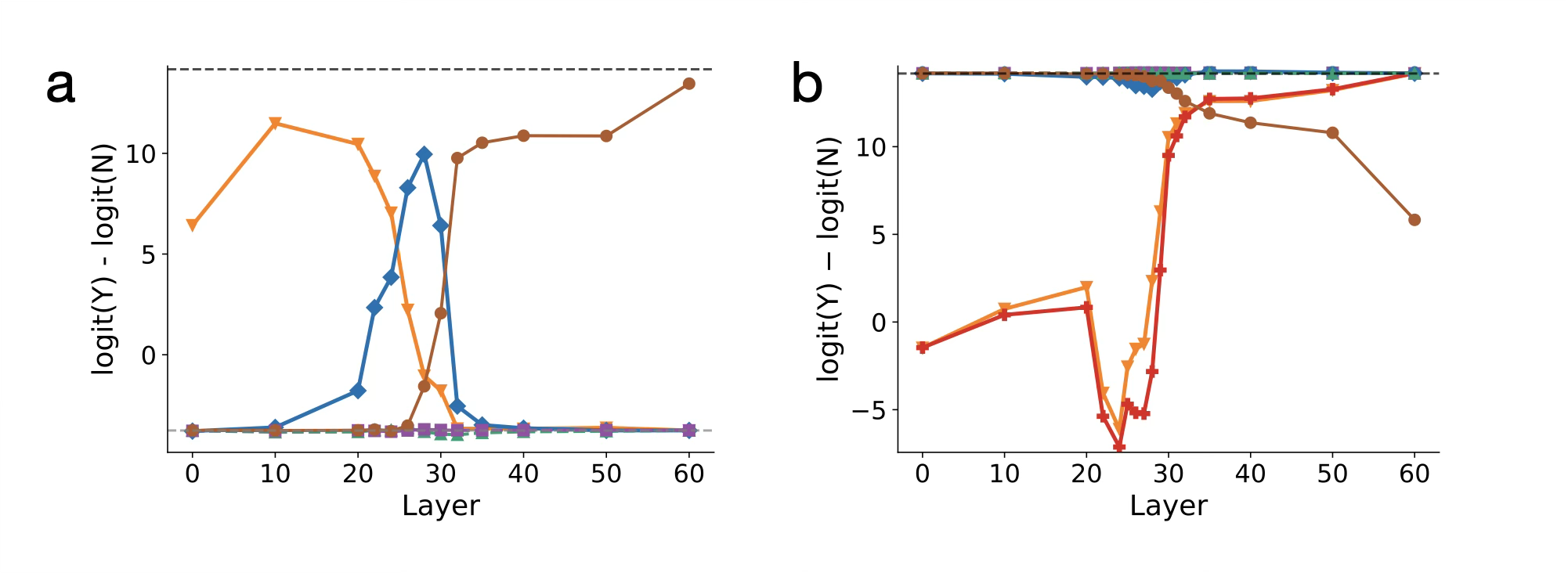}
    \caption{Gemma 3 27B causal experiments: verification logprob difference ($\log P(\text{Y}) - \log P(\text{N})$). \textbf{a}: Activation patching. Clean baseline: 14.20 (black dashed); corrupted: $-4.15$ (grey dashed). The same three positions that rescue $d'$ (Figure~\ref{fig:gemma_causal}a) also rescue the graded verification signal, with the same temporal dissociation: LAT (orange) at early layers, PANL (blue) at mid layers, prompt last token (brown) at later layers. PANL+1 and PANL offset 9 show no rescue. \textbf{b}: Activation mean ablation. LAT ablation (orange) produces large negative shifts at early-to-mid layers (peak $\Delta = -20.3$ at L24); joint LAT+PANL ablation (red) produces additional shifts at mid layers, consistent with redundancy. PANL alone (blue) shows no reliable shift. Prompt last token ablation (brown) disrupts at later layers.}
    \label{fig:gemma_causal_logdiff}
\end{figure}

\begin{figure*}[!t]
    \centering
    \includegraphics[width=\textwidth]{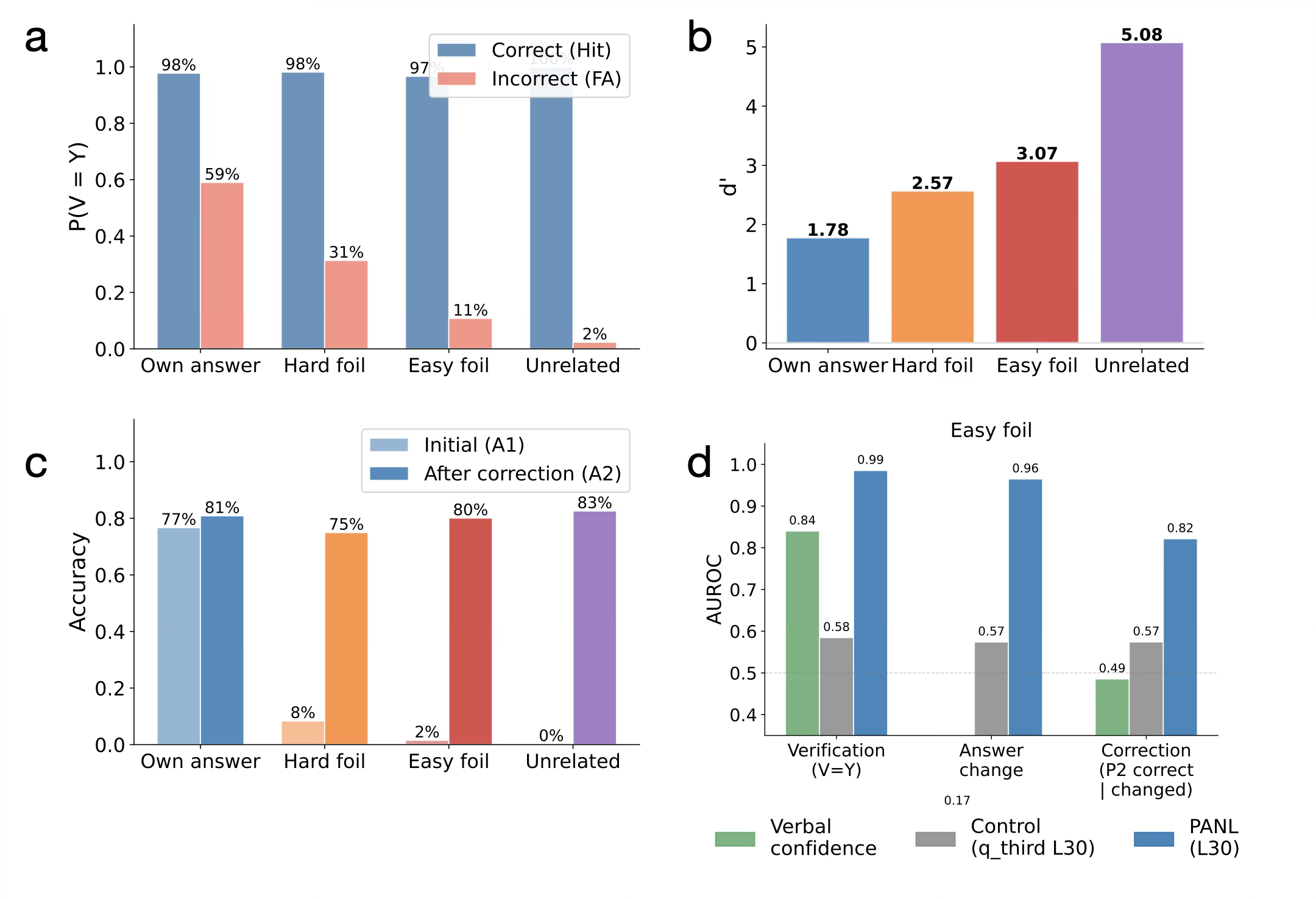}
    \caption{Foil experiment results ($n = 1{,}929$ questions per condition, same questions across all). \textbf{a}: Verification endorsement rate (V=Y) for correct answers (hit rate, blue) and incorrect answers (false alarm rate, red) by condition. Hit rates remain near ceiling ($\geq 97\%$) across all conditions, while false alarm rates decrease monotonically from own answers (59\%) through hard foils (31\%), easy foils (11\%), to unrelated foils (2\%). \textbf{b}: Error-detection sensitivity ($d'$) increases with foil discriminability, from 1.78 (own answer) to 5.08 (unrelated foil). The response criterion shifts from strong Y-bias ($c = -1.14$, own answer) toward neutrality ($c = -0.14$, unrelated), indicating that the self-affirmation bias observed in the main experiment is specific to self-generated answers. \textbf{c}: Accuracy before (A1, light) and after (A2, dark) self-correction. For foil conditions, A1 accuracy reflects the foil's incorrectness (0--8\%); after correction, accuracy rises to 75--83\%, demonstrating that the model can recover the correct answer for the majority of questions. \textbf{d}: PANL L30 probe transfer to foil conditions. Probes trained on own-answer trials predict verification (AUROC $= .84$), answer change ($.99$), and correction success ($.82$) on easy foil trials, with the control position (grey) at or below chance. As in the main experiment, verbal confidence barely predicts answer change success above chance. See Table~\ref{tab:foil_auroc_full} for results for other foils (AUROC).}
    \label{fig:foil_composite}
\end{figure*}

\clearpage

\section{Supplemental Tables}
\label{app:tables}

\begin{table*}[ht]
\centering
\caption{PANL probe score improves prediction of verification, answer change, and correction success beyond the best behavioural baseline. \emph{Behav} is an $L_2$-regularised logistic regression on all available behavioural predictors (see Tables~\ref{tab:supp_verif}--\ref{tab:supp_correct} for individual predictor AUROCs). \emph{PANL} is the cross-validated probe score from layer~30 activations at the post-answer newline position. The likelihood-ratio test assesses whether PANL adds to the combined behavioural model. Answer change and A2 correctness are restricted to incorrect trials; A2 correctness is further restricted to trials where the model changed its answer, the subset where the prediction task is non-trivial (correct trials are almost exclusively hits that do not change).}
\label{tab:AUROC_main_summary}
\resizebox{\textwidth}{!}{%
\begin{tabular}{lrcccrc}
\toprule
\textbf{Target} & \textbf{$n$} & \textbf{Behav} & \textbf{PANL} & \textbf{Behav + PANL} & \textbf{LR $\chi^2$} & \textbf{$p$} \\
\midrule
Verification Y/N (all trials)           & 7{,}223 & .908 & .986 & .983 & 1100.0 & $< .001$ \\
Answer change (incorrect)               & 1{,}764 & .901 & .921 & .931 &  237.7 & $< .001$ \\
A2 correct (incorrect + changed)        &    856  & .475 & .614 & .605 &    9.5 & $< .01$  \\
\bottomrule
\end{tabular}%
}
\end{table*}

\begin{table*}[ht]
\centering
\caption{Predicting verification response (V=Y vs V=N): individual and combined predictor AUROCs. \emph{Combined} is an $L_2$-regularised logistic regression on all behavioural predictors listed. PANL is the cross-validated probe score at layer~30. The likelihood-ratio test assesses whether PANL adds to the combined behavioural model. A1 correctness is excluded from the incorrect-only analysis (constant within subset).}
\label{tab:supp_verif}
\resizebox{\textwidth}{!}{%
\begin{tabular}{lrccccccrc}
\toprule
 & & \multicolumn{3}{c}{\textbf{Individual predictors}} & & & & & \\
\cmidrule(lr){3-5}
\textbf{Subset} & \textbf{$n$} & \textbf{Logprob} & \textbf{Conf} & \textbf{A1 corr} & \textbf{Combined} & \textbf{PANL} & \textbf{Comb + PANL} & \textbf{LR $\chi^2$} & \textbf{$p$} \\
\midrule
All trials    & 7{,}223 & .665 & .800 & .843 & .908 & .986 & .983 & 1100.0 & $< .001$ \\
Incorrect     & 1{,}764 & .530 & .703 & ---  & .715 & .958 & .948 &  443.7 & $< .001$ \\
\bottomrule
\end{tabular}%
}
\end{table*}

\begin{table*}[ht]
\centering
\caption{Predicting answer change within incorrect trials. Verification log-probability difference ($\log P(\text{Y}) - \log P(\text{N})$) is the strongest single behavioural predictor; PANL adds significant variance beyond the combined behavioural baseline.}
\label{tab:supp_change}
\resizebox{\textwidth}{!}{%
\begin{tabular}{lrcccccccrc}
\toprule
 & & \multicolumn{3}{c}{\textbf{Individual predictors}} & & & & & \\
\cmidrule(lr){3-5}
\textbf{Subset} & \textbf{$n$} & \textbf{Logprob} & \textbf{Conf} & \textbf{Verif LD} & \textbf{Combined} & \textbf{PANL} & \textbf{Comb + PANL} & \textbf{LR $\chi^2$} & \textbf{$p$} \\
\midrule
Incorrect & 1{,}764 & .508 & .657 & .907 & .901 & .921 & .931 & 237.7 & $< .001$ \\
\bottomrule
\end{tabular}%
}
\end{table*}

\begin{table*}[ht]
\centering
\caption{Predicting A2 correctness within incorrect trials. Within all incorrect trials, verification log-probability difference is a reasonable predictor (AUROC $= .740$), but within the subset that actually changed their answer---where the question is non-trivial---behavioural predictors drop to chance. PANL retains predictive value in this critical subset ($p < .01$). Note that the combined model (Comb + PANL) is slightly below PANL alone in the changed subset: with behavioural features at chance, the combined model is effectively the PANL score plus noise from uninformative covariates, which slightly degrades performance on this smaller sample ($n = 856$).}
\label{tab:supp_correct}
\resizebox{\textwidth}{!}{%
\begin{tabular}{lrcccccccrc}
\toprule
 & & \multicolumn{3}{c}{\textbf{Individual predictors}} & & & & & \\
\cmidrule(lr){3-5}
\textbf{Subset} & \textbf{$n$} & \textbf{Logprob} & \textbf{Conf} & \textbf{Verif LD} & \textbf{Combined} & \textbf{PANL} & \textbf{Comb + PANL} & \textbf{LR $\chi^2$} & \textbf{$p$} \\
\midrule
Incorrect             & 1{,}764 & .526 & .567 & .740 & .724 & .751 & .760 & 46.4 & $< .001$ \\
Incorrect + changed   &    856  & .518 & .524 & .531 & .475 & .614 & .605 &  9.5 & $< .01$  \\
\bottomrule
\end{tabular}%
}
\end{table*}

\begin{table*}[ht]
\centering
\caption{Predicting verification response, answer change, and correction success in the foil experiment. \emph{Conf}: verbal confidence alone. \emph{Verif LD}: verification log-probability difference ($\log P(\text{Y}) - \log P(\text{N})$) alone. \emph{Behav}: $L_2$-regularised logistic regression on all available behavioural predictors (confidence for verification; confidence and verification LD for answer change and correction). \emph{PANL}: cross-validated probe score at layer~30. Correction $=$ A2 correct among trials where the model changed the candidate's answer. A1 correctness is excluded from all analyses as it is constant within each foil condition (always incorrect). The likelihood-ratio test assesses whether PANL adds to the combined behavioural model.}
\label{tab:foil_auroc_full}
\resizebox{\textwidth}{!}{%
\begin{tabular}{llrccccccrc}
\toprule
 & & & \multicolumn{2}{c}{\textbf{Individual}} & & & & & & \\
\cmidrule(lr){4-5}
\textbf{Target} & \textbf{Condition} & \textbf{$n$} & \textbf{Conf} & \textbf{Verif LD} & \textbf{Behav} & \textbf{PANL} & \textbf{Behav + PANL} & \textbf{LR $\chi^2$} & \textbf{$p$} \\
\midrule
Verification Y/N 
  & Hard foil    & 1{,}924 & .886 & ---  & .879 & .981 & .977 & 442.1 & $< .001$ \\
  & Easy foil    & 1{,}924 & .840 & ---  & .827 & .985 & .983 & 258.0 & $< .001$ \\
  & Unrelated    & 1{,}929 & .775 & ---  & .771 & .995 & .994 &  84.8 & $< .001$ \\
\addlinespace
Answer change 
  & Hard foil    & 1{,}924 & .842 & .930 & .922 & .938 & .941 & 165.8 & $< .001$ \\
  & Easy foil    & 1{,}924 & .832 & .973 & .967 & .965 & .972 &  74.4 & $< .001$ \\
  & Unrelated    & 1{,}929 & .849 & .994 & .993 & .989 & .995 &  18.8 & $< .001$ \\
\addlinespace
A2 correct $|$ changed 
  & Hard foil    & 1{,}444 & .510 & .656 & .637 & .859 & .852 &  67.5 & $< .001$ \\
  & Easy foil    & 1{,}791 & .514 & .697 & .694 & .822 & .820 & 118.0 & $< .001$ \\
  & Unrelated    & 1{,}915 & .519 & .711 & .706 & .813 & .805 & 122.6 & $< .001$ \\
\bottomrule
\end{tabular}%
}
\end{table*}

\begin{table}[htbp]
\centering
\caption{Cross-task transfer of PANL L30 probes. Triv and MNLI: within-task 5-fold cross-validated AUROC. T$\to$M: TriviaQA-trained probe tested on MNLI. M$\to$T: MNLI-trained probe tested on TriviaQA. Cos: cosine similarity between probe weight vectors. Control position (question third token L30) shows no transfer on any target (all AUROCs $.45$--$.53$, all cosines $|r| < .01$).}
\label{tab:cross_transfer}
\smallskip
\resizebox{\linewidth}{!}{%
\begin{tabular}{lccccc}
\toprule
Target & Triv & MNLI & T$\to$M & M$\to$T & cos \\
\midrule
Verification Y/N & .986 & .868 & .587 & .934 & $+.132$ \\
Verif Y/N (incorrect) & .958 & .813 & .573 & .897 & $+.143$ \\
P2 correct (all) & .891 & .697 & .560 & .549 & $+.051$ \\
P2 correct (incorrect) & .752 & .811 & .627 & .719 & $+.069$ \\
P2 correct (changed) & .617 & .857 & .602 & .474 & $+.023$ \\
P2 correct (inc+changed) & .621 & .862 & .547 & .477 & $+.027$ \\
\bottomrule
\end{tabular}%
}
\end{table}

\begin{table*}[ht]
\centering
\caption{Qwen 2.5 7B: PANL probe prediction of verification and correction success. Format follows Table~\ref{tab:AUROC_main_summary}. \emph{Behav} is an $L_2$-regularised logistic regression on available behavioural predictors: answer log-probability, verbal confidence, and A1 correctness for verification (all); answer log-probability and verbal confidence for verification (errors only); answer log-probability, verbal confidence, and verification logit difference for correction success. \emph{PANL} is the cross-validated probe AUROC from layer-20 activations at the post-answer newline position.}
\label{tab:AUROC_qwen_summary}
\resizebox{\textwidth}{!}{%
\begin{tabular}{lrccc}
\toprule
\textbf{Target} & \textbf{$n$} & \textbf{Conf} & \textbf{Behav} & \textbf{PANL} \\
\midrule
Verification (all)                    & 3{,}500 & .703 & .860 & .961 \\
Verification (errors only)            & 1{,}452 & .655 & .646 & .917 \\
Correction success (incorrect)        & 1{,}245 & .461 & .719 & .774 \\
Correction success (inc.\ + changed)  &     741 & .528 & .575 & .679 \\
\bottomrule
\end{tabular}%
}
\end{table*}
\clearpage

\section{Supplemental Results: Gemma 3 27B on TriviaQA}
\label{app:gemma_suppl}

\subsection{Error Detection is Graded by Answer Plausibility}
\label{app:foil}

To test whether the model's error-detection capacity extends beyond self-generated answers, we presented the model with foil answers---wrong answers it did not produce (termed ``the candidate's answer''; see \S\ref{app:foil_method})---and asked it to verify them using the same paradigm. We compared three foil conditions, each using the same 1,929 questions: (i) a hard foil (a plausible but incorrect alternative, generated by GPT-4o), (ii) an easy foil (a topically related but clearly wrong answer), and (iii) an unrelated foil (an answer to a different question entirely).

Error-detection sensitivity increases monotonically with foil discriminability (Figure~\ref{fig:foil_composite}a,b): $d' = 2.57$ for hard foils, $3.07$ for easy foils, and $5.08$ for unrelated foils (cf.\ $d' = 1.78$ for the model's own incorrect answers). The model's evaluation is thus graded by the plausibility of the presented answer, not merely by whether the answer was self-generated.

Notably, the strong Y-bias observed in the main experiment is not a fixed property of the model's verification behaviour. The response criterion shifts toward neutrality as foil plausibility decreases: $c = -0.78$ for hard foils, $-0.33$ for easy foils, and $-0.14$ for unrelated foils (cf.\ $c = -1.14$ for own answers). False alarm rates drop accordingly---from 31\% for hard foils to 11\% for easy foils and 2\% for unrelated foils---while hit rates remain near ceiling ($\geq 97\%$) across all conditions. The Y-bias thus reflects the difficulty of the evaluation rather than a fixed tendency toward endorsement: when the presented answer is clearly wrong, the model readily rejects it.

Verbal confidence tracks this pattern: mean confidence is 0.31 for hard foils, 0.12 for easy foils, and 0.06 for unrelated foils, indicating that the model appropriately calibrates its confidence to the plausibility of the answer under evaluation. Self-correction rates mirror the verification signal: the model changes its answer on 75\% of hard foil, 93\% of easy foil, and 99\% of unrelated foil trials. Importantly, as in the main experiment, verbal confidence barely predicts answer change success above chance (see Figure~\ref{fig:foil_composite} and Table~\ref{tab:foil_auroc_full} for results for all foil types). Conditional on changing, correction success is high across all foil conditions (83--89\%), consistent with the model having generated the correct answer on its first attempt for the majority of these questions (Figure~\ref{fig:foil_composite}c).

\subsection{PANL Signal Characterisation}
\label{app:signal_char}

To characterise the nature of the evaluative signal at PANL, we examined correlations between the probe decision function score (trained to predict verification Y/N) and behavioural variables across all trials ($n = 7{,}223$). The probe score correlated most strongly with the verification logprob difference ($r = 0.91$), confirming that PANL pre-computes the verification decision before the model generates Y or N. The correlation with verbal confidence was substantial but weaker ($r = 0.60$), while ground-truth correctness ($r = 0.46$) and answer token log-probability ($r = 0.21$) showed progressively smaller relationships.

Within objectively incorrect trials ($n = 1{,}764$), the pattern sharpens. Answer log-probability becomes entirely uninformative ($r = 0.02$, $p = .46$), while the probe score continues to track the verification decision ($r = 0.83$) and verbal confidence ($r = 0.60$). The ordering---verification confidence $>$ verbal confidence $>$ correctness $>$ answer logprob---is consistent with PANL encoding a pre-computed evaluative judgment that is more closely related to the model's second-order self-evaluation than to the raw generation signal. That the probe score predicts the eventual verification output with $r = 0.91$---from activations recorded before the verification instruction---indicates that the verification decision is not computed de novo when the model encounters the verification prompt; rather, the evaluative judgment is already present in the residual stream at PANL, and the verification instruction triggers a readout of this pre-existing signal.

\subsection{P(IK) Orthogonality}
\label{app:pik}

To assess whether the PANL error-detection signal is redundant with the model's latent knowledge state, we compared the weight vectors of two linear probes trained at layer~30: one predicting the verification response (V=Y vs V=N) and one predicting P(IK), a sampling-based estimate of retrieval reliability following \citet{kadavath2022language} (20 samples at temperature $= 1$; see \S\ref{app:pik_method}). The cosine similarity between these weight vectors was $+0.007$, indicating near-complete orthogonality. The verification probe and the P(IK) probe extract geometrically independent directions from the same activation space: one encodes the model's evaluative judgment about a specific question--answer pair, while the other encodes how reliably the model can retrieve information about the topic in general. This dissociation confirms that the error-detection signal is not a proxy for question difficulty, and that PANL carries instance-specific evaluative information beyond what question-level knowledge estimates capture.

In a nested logistic regression including P(IK), verbal confidence, verification response, and mean answer log-probability as baseline predictors, adding the PANL probe decision function significantly improved prediction of both answer change (AUROC: $.919 \to .958$; LR~$\chi^2 = 101.7$, $p < 10^{-23}$) and A2 correctness (AUROC: $.915 \to .934$; LR~$\chi^2 = 99.0$, $p < 10^{-23}$). Surface text features (TF-IDF lexical features and token lengths of question and answer text) added no predictive value over the behavioural scalars ($p = 1.0$), confirming that the probe is not exploiting shallow textual cues.

\subsection{Additional Probing Analyses}
\label{app:probing_suppl}

\paragraph{Answer change and correctability within FA and CR subsets.}
Within false alarm trials---where the model endorsed its incorrect answer yet still changed it 26.8\% of the time---PANL at layer~30 predicts answer change with AUROC $= .867$ ($\Delta = +.311$ above the behavioural baseline of $.556$), substantially outperforming the last answer token ($\Delta = +.239$) and the control position ($\Delta = -.004$). This indicates that even when the model's overt verification response signals endorsement, PANL distinguishes trials where residual uncertainty will drive revision from those where the model fully commits to its error. Within correct rejection trials ($n = 540$, of which 528 changed), the near-ceiling change rate leaves almost no variance to predict. PANL also predicts A2 correctness within both FA and CR subsets above the behavioural baseline, indicating that correctability information is present regardless of whether the model overtly detected the error.

\paragraph{Continuous prediction of verification strength.}
Beyond the binary verification decision, PANL activations predict the verification logprob difference ($\log P(\text{Y}) - \log P(\text{N})$) as a continuous variable. Using Ridge regression, the Pearson correlation between PANL probe predictions and verification logprob difference increases sharply at mid-layers, with PANL leading other positions. This indicates that PANL does not merely encode which side of a decision boundary the model will fall on, but tracks the \emph{strength} of the verification decision---consistent with a graded confidence representation rather than a binary error flag.

\paragraph{Information propagation to downstream positions.}
Downstream token positions (offset 6, 12, and 18 tokens after PANL, corresponding to positions within the verification instruction) also carry verification-predictive information at later layers, consistent with evaluative information propagating forward from PANL through the residual stream. However, as shown in the causal analyses (\S\ref{app:causal_detail}), only PANL plays a causal role in error detection---decodable information at downstream positions does not contribute causally to the verification decision.

\subsection{Causal Experiment Details}
\label{app:causal_detail}

\paragraph{Activation patching: full results.}
Figure~\ref{fig:gemma_causal}A shows $d'$ as a function of patching layer for five token positions. Patching LAT rescues error detection at early-to-mid layers (peak $d' = 1.23$ at L15, $\sim$100\% recovery), declining sharply after L30. Patching PANL rescues at mid layers (peak $d' = 0.89$ at L30, $\sim$74\% recovery), with a complementary temporal profile---rising as LAT declines. Patching the prompt last token rescues at later layers (peak $d' = 1.25$ at L35, $\sim$107\% recovery), consistent with this position assembling the final verification decision from information propagated forward through the residual stream. Patching PANL+1 and PANL offset~9---positions where linear probes successfully decode verification-predictive information---has no effect on $d'$ at any layer, remaining at the corrupted baseline throughout. This dissociation between decodable information and causal influence is consistent with prior findings that linearly decodable signals are not always expressed in downstream behaviour \citep{azaria2023internal}. The verification logprob difference ($\log P(\text{Y}) - \log P(\text{N})$) shows the same pattern (Figure~\ref{fig:gemma_causal_logdiff}), confirming that patching affects not only the binary verification decision but also its graded strength.

\paragraph{Activation noising (through mean ablation): full results.}
Ablating PANL alone has no detectable effect on $d'$ at any layer (maximum $|\Delta d'| = 0.062$ at L27), with error-detection sensitivity remaining at the clean baseline ($d' = 1.20$) throughout. Ablating LAT severely disrupts error detection at early-to-mid layers ($d' = -0.14$ at L0; $-0.18$ at L10; $-0.21$ at L22), with gradual recovery from L25 ($d' = 0.44$) reaching baseline by L30 ($d' = 1.26$). Ablating the prompt last token produces no effect at early-to-mid layers but severe disruption at later layers ($d' = 0.0$ at L40--60). Ablating PANL+1 and PANL offset~9 has no effect at any layer (all $|\Delta d'| < 0.03$). The verification logprob difference shows the same pattern (Figure~\ref{fig:gemma_causal_logdiff}): LAT ablation produces large negative shifts (e.g., $\Delta = -20.3$ at L24) at early-to-mid layers, prompt last token ablation at later layers ($\Delta = -8.3$ at L60), and PANL ablation produces no reliable shift (maximum $|\Delta| = 0.90$).

\paragraph{Joint ablation details.}
Jointly ablating LAT and PANL produces deficits exceeding LAT ablation alone at mid layers. At L27, LAT alone yields $d' = 0.82$ ($\Delta\text{logit\_diff} = -15.4$) while joint ablation yields $d' = 0.31$ ($\Delta\text{logit\_diff} = -19.4$). At L28, LAT alone yields $d' = 1.05$ ($\Delta\text{logit\_diff} = -11.8$) while joint ablation yields $d' = 0.58$ ($\Delta\text{logit\_diff} = -17.0$). The effect is concentrated at mid layers (25--29) and dissipates by L30.

\paragraph{LAT confound.}
LAT serves a dual role: it carries both the answer representation itself and any evaluative signal computed over that content. Corrupting the answer token necessarily corrupts both, making it impossible to determine from LAT manipulations alone whether disruption reflects loss of the answer representation, loss of the evaluative signal, or both. PANL provides a cleaner window onto the evaluative computation: it is not part of the answer itself, yet carries a rich evaluative signal that is causally sufficient to restore error detection. For this reason, probing analyses focus on PANL as the primary locus of the evaluative representation, while acknowledging that the causal evidence is consistent with the evaluative signal being distributed across both positions.

\clearpage

\section{Cross-Model and Cross-Task Generalisation}
\label{app:generalisation}

\subsection{MNLI Behavioural Results}
\label{app:mnli_behav}

Gemma 3 27B scored 71.6\% on the MNLI dataset at first attempt (A1: $n = 9{,}888$). Self-correction yielded a modest but significant improvement to 73.9\% at A2 ($\Delta = +2.3$ percentage points; McNemar's $\chi^2 = 50.6$, $p < .001$) --- though there was a marked improvement within neutral trials (see below). Across all labels, the model displayed weak error-detection sensitivity ($d' = 0.70$).

Answer change rates varied dramatically by label: the model changed its answer on only 0.8\% of entailment trials and 3.3\% of contradiction trials, compared to 30.3\% of neutral trials. We therefore restrict all subsequent MNLI analyses to neutral trials ($n = 3{,}395$), where the model displays meaningful self-correction behaviour.

\begin{figure}[!t]
    \centering
    \includegraphics[width=\linewidth]{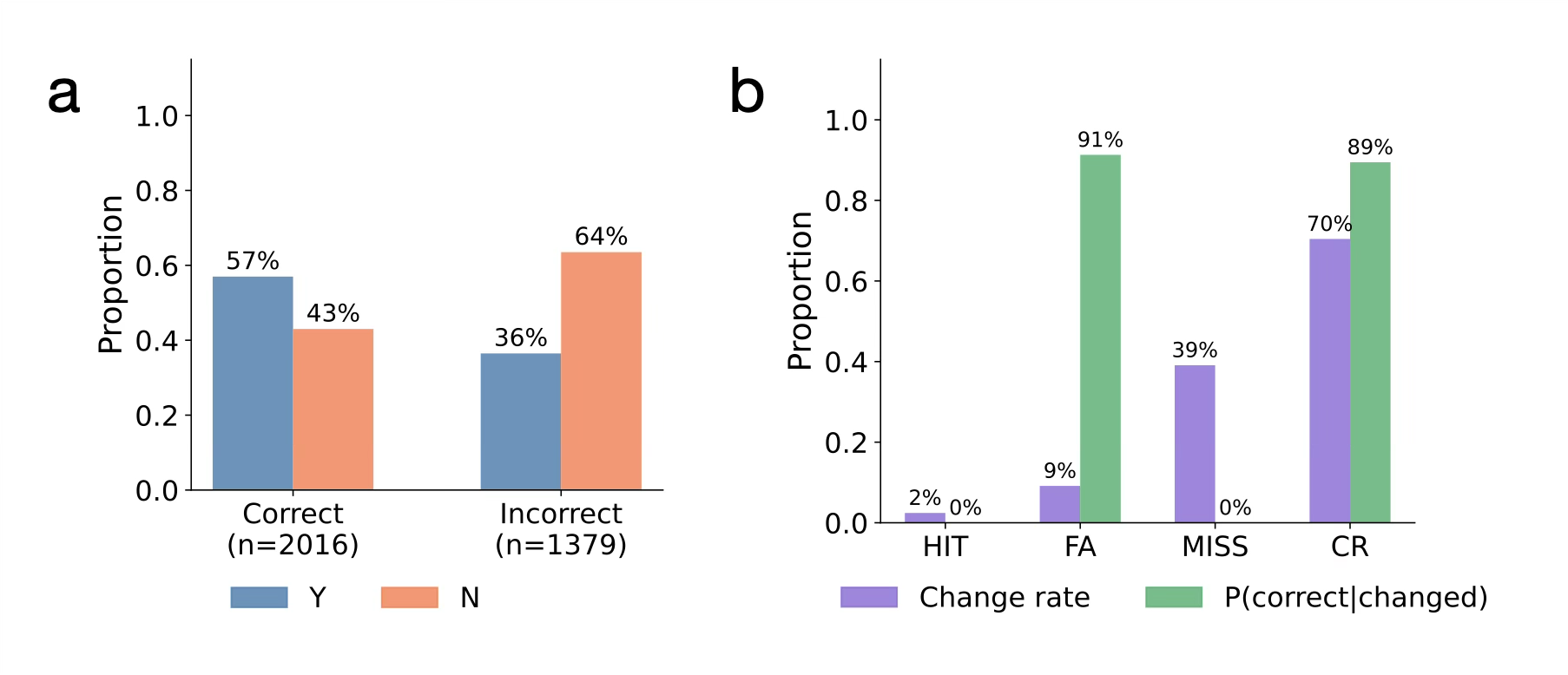}
    \caption{MNLI neutral behavioural results ($n = 3{,}395$). \textbf{a}: Verification responses by A1 correctness ($d' = 0.53$, $c = 0.15$). Unlike TriviaQA, the model shows a near-neutral criterion. \textbf{b}: Answer change rate and correction success by SDT cell. Verification gates revision as in TriviaQA, but correction success is concentrated in correct rejections (89.6\%); misses yield 0\% correction by construction (changing away from the correct label is always wrong in 3-way classification).}
    \label{fig:mnli_behav_composite}
\end{figure}

\paragraph{Verification behaviour on neutral trials.}
Figure~\ref{fig:mnli_behav_composite} shows the distribution of verification responses as a function of A1 correctness for neutral trials. The model detects its errors at a modest rate ($d' = 0.53$, $c = 0.15$), with a near-neutral criterion---markedly different from the strong Y-bias observed in TriviaQA ($c = -1.34$). When A1 is correct, the model confirms its answer 54.7\% of the time; when A1 is incorrect, the model responds V=Y on 33.9\% of trials. Self-correction on neutral trials yields a marked improvement in accuracy from 59.4\% to 66.1\% ($\Delta = +6.7$ percentage points; McNemar's $\chi^2 = 53.1$, $p < .001$).

\paragraph{Verification determines whether the model revises, but not whether revision succeeds.}
Figure~\ref{fig:mnli_behav_composite} shows answer change rates and correction success by SDT cell for neutral trials. As in TriviaQA, the verification response gates revision: correct rejections show a 72.7\% change rate compared to 0.2\% in false alarms, and misses show 40.1\% compared to 0.1\% in hits. Conditional on changing, correction success is high in correct rejections (89.6\%) but by nature near zero in misses (0.0\%), a pattern that differs from TriviaQA's roughly constant $\sim$30--34\% correction rate across cells. This difference likely reflects the structure of the task: in MNLI, switching from neutral to the correct alternative (entailment or contradiction) requires identifying a specific label, and the model appears substantially better at this when it has correctly detected an error (CR).

\paragraph{Confidence drives the verification decision but is a weaker signal than in TriviaQA.}
Verbal confidence carries evaluative information beyond log-probabilities on neutral trials, replicating the second-order behavioural signature observed in TriviaQA. Adding verbal confidence to a logistic model containing only the mean answer log-probability improves prediction of the verification response (LR~$\chi^2 = 13.5$, $p < .001$; $\beta_{\mathrm{conf}} = +0.069$, $\beta_{\mathrm{logprob}} = +0.003$). Confidence also adds beyond log-probability and A1 correctness combined (LR~$\chi^2 = 17.2$, $p < .001$). Within incorrect trials, confidence predicts verification (AUROC $= .567$, $\chi^2 = 9.3$, $p = .002$). However, these effects are substantially weaker than in TriviaQA ($\beta_{\mathrm{conf}} = 0.069$ vs $0.51$; within-incorrect AUROC $= .567$ vs $.737$), suggesting that the model's evaluative signal for NLI is less robust than for factual QA.

Verbal confidence correlates weakly with the verification logprob difference ($r = .090$, $p < .001$; within correct $r = .086$; within incorrect $r = .145$), a much weaker association than in TriviaQA ($r = .604$). Mean answer log-probability is essentially unrelated to verification ($r = -.022$, $p = .20$), confirming that generation-time fluency provides no useful signal for NLI verification. Unlike TriviaQA, where verbal confidence followed a clean SDT-consistent ordering (HIT $>$ FA $>$ MISS $>$ CR), the pattern in MNLI neutral shows: FA (0.614) $>$ HIT (0.566). The reversal of HIT and FA is notable: the model expresses higher confidence on incorrect trials it endorses than on correct trials it endorses, the opposite of TriviaQA. The compressed overall confidence range further indicates that the model's confidence signal carries less information about its own accuracy on this task.

\paragraph{Verbal confidence or verification confidence does not predict correction quality.}
As in TriviaQA, verbal confidence does not predict which revisions succeed. Within all changed trials, confidence is uninformative about A2 correctness (AUROC $= .529$, $r = +.056$, $p = .072$, $n = 1{,}030$). Within incorrect changed trials, the same null holds (AUROC $= .527$, $r = -.035$, $p = .37$, $n = 663$). The verification logprob difference similarly fails to predict correction success within changed trials (AUROC $= .565$, $r = -.115$, $p < .001$, $n = 1{,}030$; within incorrect changed: AUROC $= .552$, $r = -.058$, $p = .14$, $n = 663$). Both signals determine whether the model revises but not whether revision succeeds---replicating the TriviaQA dissociation and motivating the PANL probing analyses that follow.

\subsection{MNLI: PANL Activations Predict Error Detection and Self-Correction}
\label{sec:mnli_probing}

We asked whether the PANL evaluative signal identified in TriviaQA generalises to a structurally different task. MNLI neutral trials present a distinct challenge: the model's behavioural confidence is weakly informative (AUROC $= .534$ for verification prediction), the confidence range is compressed, and the SDT-cell ordering does not follow the clean pattern observed in TriviaQA. If PANL encodes a domain-general evaluative signal, it should nonetheless predict verification and correction outcomes beyond behavioural baselines.

\paragraph{PANL predicts verification responses beyond behavioural signals.}
PANL activations at layer 30 predict verification responses with AUROC $= .868$, far exceeding verbal confidence alone (AUROC $= .534$) and the control position (third question token: AUROC $= .503$). Within incorrect trials, PANL achieves AUROC $= .813$ (control: $.499$). The gap between PANL and behavioural baselines is even larger here than in TriviaQA, because the behavioural signals are so weak on this task---the model's overt confidence captures very little about its own verification behaviour on MNLI neutral, yet the internal representation at PANL carries a strong signal.

\paragraph{PANL predicts answer changes.}
PANL at layer 30 predicts whether the model will change its answer with AUROC $= .869$, exceeding the combined behavioural baseline of verbal confidence, A1 correctness, and verification response ($\Delta = +.045$ above baseline AUROC $= .823$). The control position shows no improvement (AUROC $= .523$). Within incorrect trials, PANL achieves AUROC $= .811$ ($\Delta = +.024$ above baseline $.787$).

\paragraph{PANL predicts self-correction success where behavioural signals fail.}
Replicating the central TriviaQA finding, PANL predicts which revisions will succeed where behavioural signals cannot. Across all trials, PANL predicts A2 correctness with AUROC $= .697$ ($\Delta = +.143$ above baseline $.555$; control: $.541$). The effect is strongest in the subsets where behavioural signals are at chance. Within trials where the model changed its answer---where verbal confidence is uninformative about correction success (AUROC $= .529$)---PANL achieves AUROC $= .857$ ($\Delta = +.335$ above baseline $.522$). Within incorrect trials that changed---the subset where self-correction matters most, and where the combined behavioural baseline is at chance ($.512$)---PANL achieves AUROC $= .862$ ($\Delta = +.351$). The control position shows no predictive power in any of these subsets ($\leq .543$).

The correctability prediction is numerically stronger on MNLI neutral than on TriviaQA (AUROC $= .862$ vs $.738$ within incorrect trials), likely reflecting the higher base rate of successful correction on this task (89.6\% in CR trials vs $\sim$34\% in TriviaQA). Nonetheless, the qualitative pattern replicates: PANL encodes a graded signal about whether the model can produce a correct second-pass answer, and this signal is not reducible to any combination of behavioural confidence measures.

\paragraph{Summary.}
The PANL evaluative signal generalises from factual QA to natural language inference. Despite substantially weaker behavioural signals, PANL activations at layer 30 predict verification, answer change, and---most critically---correction success on MNLI neutral trials, replicating all key findings from TriviaQA. The model's internal evaluative representation is richer than what its behavioural signals convey, across both tasks.

\subsection{Cross-Task Transfer}
\label{sec:transfer}

To test whether the evaluative signal occupies the same representational subspace across tasks, we trained PANL L30 probes on one task and applied them to the other (Table~\ref{tab:cross_transfer}).

Transfer is asymmetric and target-dependent. For verification prediction, the MNLI-trained probe transfers well to TriviaQA (AUROC $= .934$, vs $.986$ within-task) but the reverse transfer is near chance ($.587$, vs $.868$ within-task). The same asymmetry holds within incorrect trials (M$\to$T: $.897$; T$\to$M: $.573$). This suggests that the MNLI verification direction captures a general evaluative feature that is also present in TriviaQA's richer representational space, while TriviaQA's verification direction is too task-specific to generalise.

For correctability prediction, the picture is mixed. Within all incorrect trials, moderate partial transfer emerges in both directions (T$\to$M: $.627$; M$\to$T: $.719$; cosine $= +.069$). However, within the more targeted subsets---trials where the model actually changed its answer---transfer collapses (changed: T$\to$M $= .602$, M$\to$T $= .474$; incorrect + changed: T$\to$M $= .547$, M$\to$T $= .477$; cosine $\leq .027$). Cosine similarity between probe weight vectors is correspondingly low ($\leq .027$ for correctability targets vs $+.132$ for verification). The control position (third question token) shows no within-task predictive power and no transfer on any target.

The evaluative signal thus generalises \emph{functionally}---PANL predicts verification and correctability on both tasks---but the linear directions encoding these signals are largely task-specific. The verification direction shows partial overlap, consistent with a general error-detection signal that both tasks share. The correctability direction does not, consistent with this signal reflecting task-specific knowledge structures: knowing whether a wrong trivia answer can be corrected draws on different knowledge than knowing whether a misclassified NLI pair can be relabelled.

\subsection{Qwen 2.5 7B Behavioural Results}
\label{app:qwen_behav}

Qwen 2.5 7B scored 58.5\% on the TriviaQA dataset at first attempt (A1: $n = 3{,}500$). Self-correction yielded a significant improvement to 63.2\% at A2 ($\Delta = +4.7$ percentage points; McNemar's $\chi^2 = 106.3$, $p < .001$), a larger gain than observed in Gemma 3 27B ($+3.7$ pp).

\begin{figure}[!t]
    \centering
    \includegraphics[width=\linewidth]{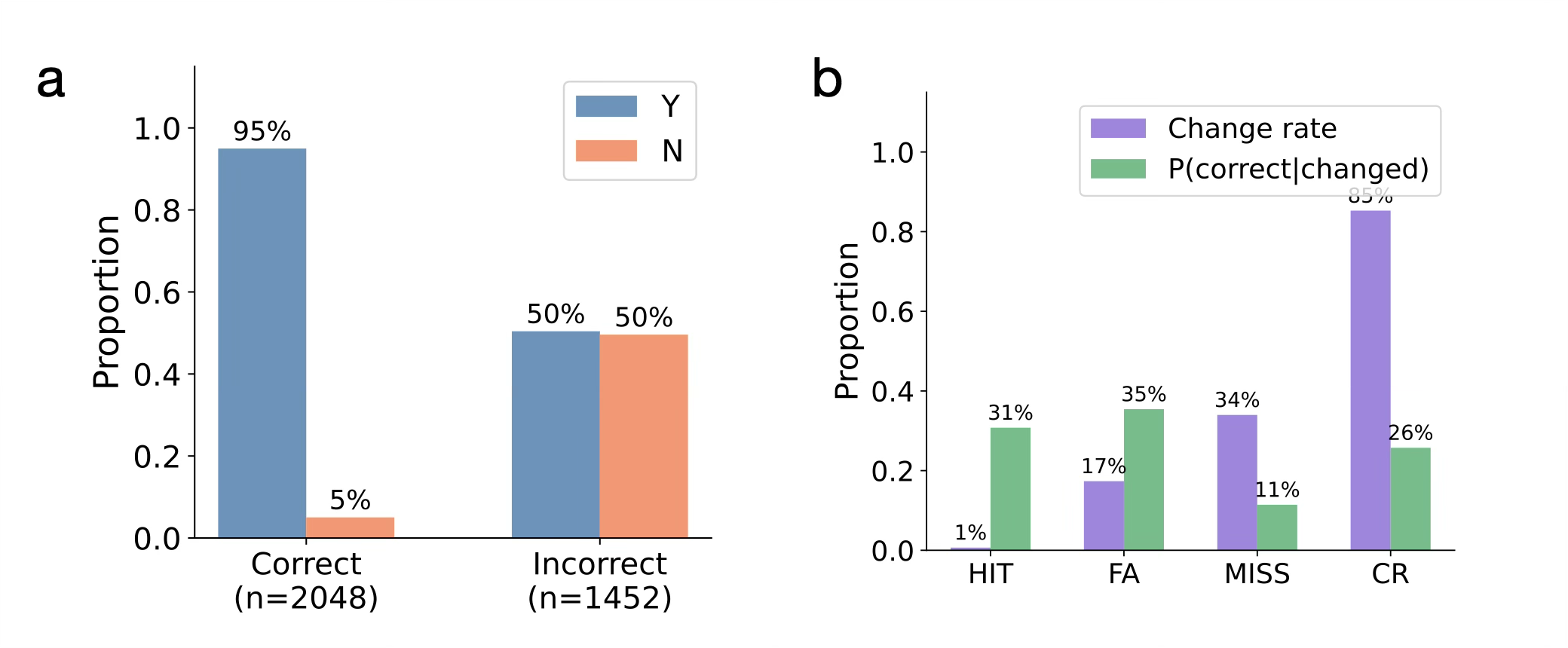}
    \caption{Qwen 2.5 7B verification and self-correction results on TriviaQA ($n = 3{,}500$). \textbf{a}: Verification responses by A1 correctness ($d' = 1.63$, $c = -0.83$). \textbf{b}: Answer change rate (purple) and correction success conditional on changing (green) by SDT cell. The pattern replicates Gemma 3 27B: verification gates revision, but correction quality is roughly constant across cells.}
    \label{fig:qwen_behavioral}
\end{figure}

\paragraph{Verification behaviour.}
The model displays robust error-detection sensitivity ($d' = 1.63$, $c = -0.83$), comparable to Gemma ($d' = 1.67$, $c = -1.34$) though with a less extreme Y-bias (see Figure~\ref{fig:qwen_behavioral}). When A1 is correct, the model confirms its answer 95.0\% of the time; when A1 is incorrect, the model responds V=Y on 50.4\% of trials, detecting its error on approximately half of incorrect trials --- a substantially higher detection rate than Gemma (30.8\%).

\paragraph{Verbal confidence carries predictive information beyond log-probabilities.}
As in Gemma, verbal confidence alone predicts verification (with AUROC $= .703$); answer log-probability alone achieves AUROC $= .750$; the combined model yields AUROC $= .812$. Within incorrect trials, verbal confidence achieves AUROC $= .655$ while answer log-probability drops to AUROC $= .534$, consistent with the pattern observed in Gemma. The correlation between verbal confidence and the verification logprob difference is $r = .535$ (within correct $r = .371$; within incorrect $r = .457$).

\paragraph{Verification determines whether the model revises, but not whether revision succeeds.}
The verification response gates revision as in Gemma: correct rejections show an 85.3\% change rate compared to 17.3\% in false alarms, and misses show 34.0\% compared to 0.7\% in hits (see Figure~\ref{fig:qwen_behavioral}).

As in Gemma, conditional on changing, correction quality is roughly constant across cells: 30.8\% in hits ($n = 13$, interpret cautiously), 35.4\% in false alarms, 11.4\% in misses, and 25.7\% in correct rejections. The overall correction rate among changed incorrect trials is 27.4\% (see Figure~\ref{fig:qwen_behavioral}).

\paragraph{Neither confidence signal predicts correction quality.}
As in Gemma, neither verbal confidence nor the verification logprob difference predicts which revisions succeed. Within incorrect changed trials, verbal confidence is uninformative about A2 correctness (AUROC $= .528$, $r = +.042$, $p = .25$, $n = 741$). The verification logprob difference shows the same null (AUROC $= .506$, $r = -.019$, $p = .61$). Both signals determine whether the model revises but not whether that revision succeeds, replicating the dissociation observed in Gemma and motivating the PANL probing analyses.

\paragraph{Summary.}
Qwen 2.5 7B replicates the key behavioural findings from Gemma 3 27B despite being a substantially smaller model (7B vs 27B) with lower baseline accuracy (58.5\% vs 75.5\%). Error detection is robust ($d' = 1.63$), verbal confidence carries evaluative information beyond log-probabilities, and confidence predicts whether errors are detected but not whether they can be corrected. The self-correction improvement is slightly larger than Gemma ($+4.7$ vs $+3.7$ pp), and the model detects a higher proportion of its errors (49.6\% vs 30.8\%), though with a lower correction success rate when it does change (27.4\% vs $\sim$34\%).

\subsection{Qwen 2.5 7B: PANL Activations Predict Error Detection and Self-Correction}
\label{sec:qwen_probing}

\begin{figure}[!t]
    \centering
    \includegraphics[width=\linewidth]{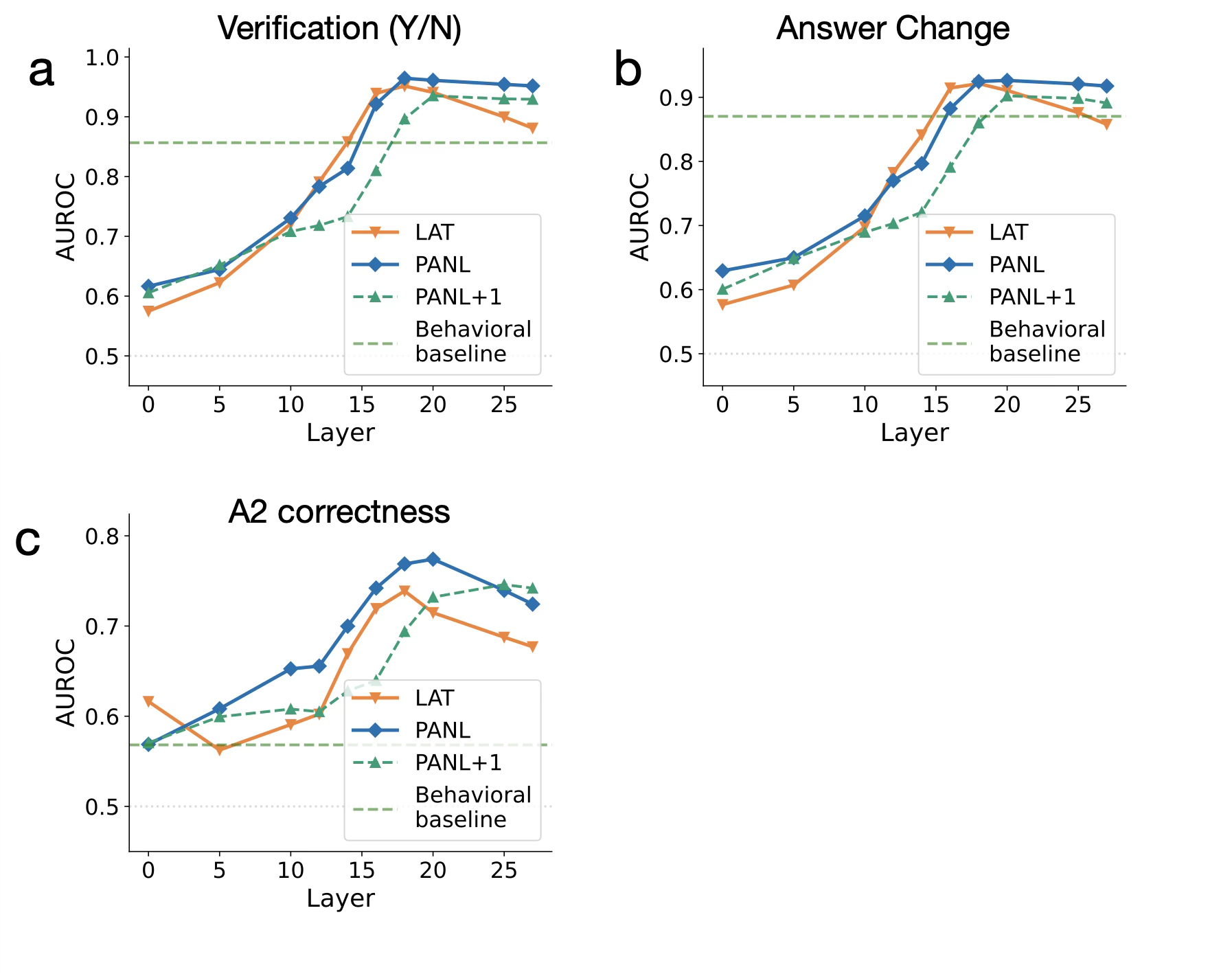}
    \caption{Qwen 2.5 7B linear probe results on TriviaQA ($n = 3{,}500$). AUROC for predicting verification response, answer change, and A2 correctness from residual stream activations at three token positions: LAT (orange), PANL (blue), and PANL+1 (green). Dashed lines: behavioural baseline (A1 correctness + mean answer log-probability + verbal confidence). \textbf{a}: Verification response (V=Y vs V=N). \textbf{b}: Answer change. \textbf{c}: A2 correctness within A1 incorrect trials. All three targets show probing performance exceeding the behavioural baseline at mid-to-upper layers, with LAT and PANL performing comparably --- replicating the Gemma 3 27B pattern in a smaller model with a shallower layer hierarchy (28 layers vs 62).}
    \label{fig:qwen_probing}
\end{figure}

We tested whether the evaluative signal replicates in Qwen 2.5 7B, a smaller model with 28 layers. Figure~\ref{fig:qwen_probing} and Table~\ref{tab:AUROC_qwen_summary} summarise the probing results.

The key findings replicate: PANL predicts verification with AUROC $= .961$ ($\Delta = +.101$ above the behavioural baseline), and within errors only achieves $.917$ where the behavioural baseline drops to $.646$. Critically, PANL predicts correction success within incorrect trials (AUROC $= .774$, $\Delta = +.055$) and within incorrect changed trials ($.679$, $\Delta = +.104$), where behavioural predictors are near chance (Table~\ref{tab:AUROC_qwen_summary}). PANL also tracks the graded strength of the verification decision ($r = .909$ with the verification logprob difference). The evaluative representation at PANL thus generalises across model families and scales, replicating in a 7B model despite lower baseline accuracy (58.5\% vs 75.5\%) and a shallower layer hierarchy.

\subsection{Qwen 2.5 7B: Causal Experiments}
\label{sec:qwen_causal}

\begin{figure}[!t]
    \centering
    \includegraphics[width=\linewidth]{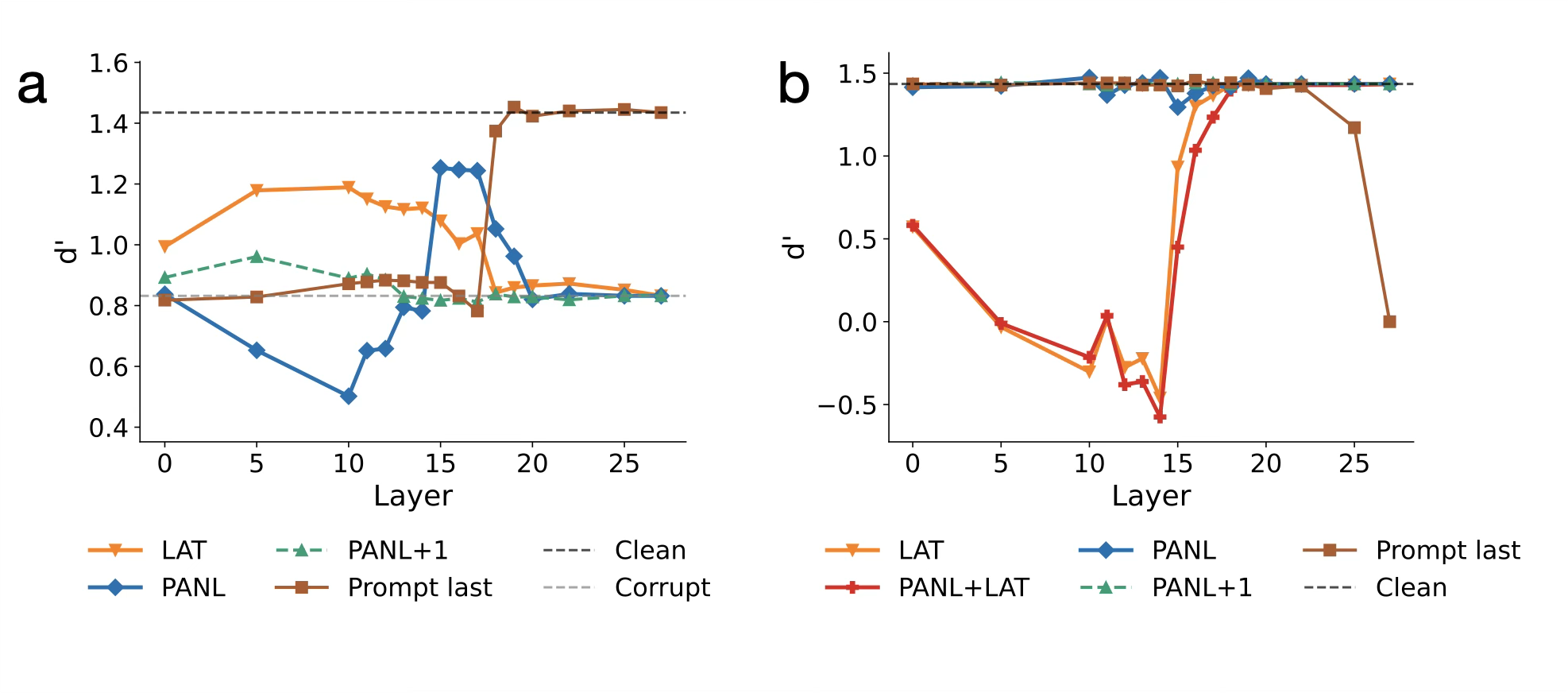}
    \caption{Qwen 2.5 7B causal experiments on TriviaQA. \textbf{a}: Activation patching. Clean $d' = 1.44$; corrupted $d' = 0.83$ (grey dashed; the above-chance corrupt baseline likely reflects question-difficulty information carried by the question tokens, which remain intact). LAT (orange) rescues at early layers (peak $d' = 1.19$ at L10, $\sim$59\% recovery); PANL (blue) rescues at mid layers (peak $d' = 1.25$ at L15, $\sim$70\% recovery); prompt last token (brown) rescues at later layers (peak $d' = 1.45$ at L19, $\sim$103\% recovery). PANL+1 (green dashed) shows no rescue. \textbf{b}: Activation noising experiment with mean ablation. Ablating LAT (orange) severely disrupts error detection at early-to-mid layers ($d' = -0.30$ at L10), recovering by L16. Ablating PANL alone (blue) has no effect (all $|\Delta d'| \leq 0.14$). Jointly ablating LAT and PANL (red) produces additional deficits beyond LAT alone at the critical mid-layer transition: at L15, LAT alone yields $d' = 0.93$ while the joint ablation yields $d' = 0.45$ ($\Delta = -0.48$), consistent with representational redundancy.}
    \label{fig:qwen_causal}
\end{figure}

We replicated the causal experiments on Qwen 2.5 7B (Figure~\ref{fig:qwen_causal}). Corruption of answer tokens reduced $d'$ from 1.44 to 0.83 --- a weaker disruption than in Gemma ($d' = 0.09$), likely because question tokens alone carry sufficient difficulty information to support partial error detection in this model.

Activation patching reveals the same three-position pattern: LAT rescues at early layers (peak $d' = 1.19$ at L10, $\sim$59\% recovery), PANL at mid layers (peak $d' = 1.25$ at L15, $\sim$70\% recovery), and the prompt last token at later layers (peak $d' = 1.45$ at L19, full recovery). PANL+1 shows no rescue at any layer. PANL is thus causally sufficient for error detection in Qwen, as in Gemma.

Mean ablation confirms the complementary pattern: ablating LAT severely disrupts error detection at early-to-mid layers ($d' = -0.30$ at L10), with recovery by L16. Ablating PANL alone has no effect ($|\Delta d'| \leq 0.14$). Jointly ablating LAT and PANL produces an additional deficit at the critical transition layers where LAT is recovering: at L15, LAT alone yields $d' = 0.93$ while the joint ablation yields $d' = 0.45$. As in Gemma, the additional PANL deficit is visible only when LAT is also disrupted, consistent with the two positions carrying the evaluative signal redundantly.

The causal architecture thus replicates across model families: PANL is sufficient but not necessary, LAT is necessary at early layers, and both positions carry the evaluative signal redundantly at mid layers.

\end{document}